\documentclass{article} % For LaTeX2e
\usepackage{iclr2027_conference,times}

\usepackage{amsmath,amsfonts,bm}

\def\eqref#1{equation~\ref{#1}}
\def\1{\bm{1}}

\DeclareMathAlphabet{\mathsfit}{\encodingdefault}{\sfdefault}{m}{sl}
\SetMathAlphabet{\mathsfit}{bold}{\encodingdefault}{\sfdefault}{bx}{n}

\usepackage{graphicx}
\usepackage{subcaption}
\usepackage{float}
\usepackage{wrapfig}
\usepackage{tabularray}
\usepackage{dsfont}
\UseTblrLibrary{booktabs}
\usepackage{rotating}
\usepackage{multirow} 
\usepackage{graphicx}
\usepackage[normalem]{ulem}
\usepackage{amsmath,amsfonts,bm}
\usepackage{tabularx}
\usepackage{pdflscape}
\usepackage{ulem}
\usepackage{bm}

\usepackage{xcolor}     
\usepackage[table]{xcolor}
\usepackage{adjustbox}  
\usepackage{amsmath}
\usepackage{enumitem}
\usepackage{hyperref}
\usepackage{url}
\usepackage{fancyvrb}
\usepackage{tcolorbox}
\tcbuselibrary{listings,breakable,skins}
\definecolor{PostDiagPromptBlue}{HTML}{3671B2}
\definecolor{PostDiagPromptFill}{HTML}{F3F7FD}
\newtcblisting{PostDiagPrompt}[1][]{
  enhanced,breakable,listing only,
  colback=PostDiagPromptFill,colframe=PostDiagPromptBlue,
  boxrule=0.6pt,arc=3mm,
  left=8pt,right=8pt,top=4pt,bottom=4pt,
  before skip=5pt,after skip=6pt,
  colbacktitle=PostDiagPromptFill,coltitle=PostDiagPromptBlue,
  fonttitle=\normalfont\small\itshape,
  listing options={basicstyle=\ttfamily\fontsize{8.5}{9.0}\selectfont,
    columns=fullflexible,keepspaces=true,breaklines=true,
    breakatwhitespace=true,breakindent=0pt,showstringspaces=false,
    aboveskip=0pt,belowskip=0pt},#1
}

\title{From Score to Diagnosis: Search-Independent Post-Diagnosis for Symbolic Regression}

\author{Qian Li$^{1,2}$, Yuxiao Hu$^{2,3}$, Juncheng Liu$^{4}$, Yanhao Jia$^{5}$, Yuntian Chen$^{2}$\thanks{Corresponding author} \\
$^{1}$Shanghai Jiao Tong University, Shanghai, China\\
$^{2}$Eastern Institute of Technology, Ningbo, China\\
$^3$The Hong Kong Polytechnic University, Hong Kong, China\\
$^4$Imperial College London, London, England\\
$^5$Nanyang Technological University, Singapore\\
\texttt{qianl01205@sjtu.edu.cn}, \texttt{yuxiao.hu@connect.polyu.hk}\\
\And
}

\iclrfinalcopy %
\begin{document}

\maketitle
\lhead{Preprint}

\begin{abstract}
Symbolic regression (SR) solvers employ diverse search mechanisms, yet most rely on the same low-information scalar fitness. Such scores provide little guidance on why a skeleton fails or how it should be revised. Recent methods develop new methods to address this bottleneck. Their enhanced feedback mechanisms are tightly coupled to their search processes and are difficult to reuse across diverse SR solvers. Instead, we ask whether a shared feedback mechanism can further improve diverse solvers facing the same bottleneck. To this end, we propose PostDiag, a search-independent feedback mechanism, which can be shared to different SR solvers. PostDiag introduces Patch-wise Feature Profiling (PFP) to capture local behavior through patch-wise features of level, trend, and roughness. Based on PFP, PostDiag quantifies Skeleton Expressiveness Risk by comparing the fitted candidate with the observations, and Skeleton Stability Risk by measuring variability of behavior after bootstrap refitting. The Diagnostic Risk Plane (DRP) combines two risks to determine whether a skeleton should be preserved, repaired, simplified, or rewritten, while the patch-wise discrepancies provide evidence for the intervention. An LLM editor then performs the diagnosis-guided intervention, and the validated revision is returned to the original solver through a lightweight adapter. Experiments across multiple LLM-based and GP-based SR solvers show that PostDiag consistently improves accuracy and accelerates convergence.

\end{abstract}

\section{Introduction}
Symbolic regression (SR) discovers compact, interpretable mathematical expressions from observed data, enabling data-driven scientific discovery across physics, biology, chemistry, and engineering~\citep{wang2023scientific,zhang2025artificial,karniadakis2021physics,krenn2022scientific,udrescu2020ai,udrescu2020ai2}. Given observations $\{(\bm{x}_i,y_i)\}_{i=1}^{N}$, SR searches for an expression $f:\bm{x}\mapsto y$ that explains the underlying relationship. Modern SR solvers employ diverse search strategies to enable efficient and accurate SR: PySR uses multi-population evolutionary search~\citep{pysr}; LLM-SR iteratively refines equations using the code-generation and reasoning capabilities of LLMs~\citep{llmsr}; LaSR co-evolves equations with a reusable concept library~\citep{lasr}; and SR-Scientist autonomously integrates data analysis, tool use, and equation refinement~\citep{xia2025sr}. This diversity is valuable because different solvers offer complementary strengths.

Despite their diverse search strategies, most SR solvers rely on the same impoverished search feedback: a scalar fitness score, such as MSE or NMSE. By collapsing the residual landscape into a single value, these scores obscure error geometry and reveal little about where, why, or how a candidate fails.Indeed, similar scores may correspond to fundamentally distinct failure modes~\citep{gensr,xiangmod}. This limited feedback leaves even sophisticated search mechanisms poorly informed, resulting in inefficient and misdirected exploration.

Recent methods RESTART and Deliberate Evolution (DE) design new SR solvers that embed structured diagnostics into their search loops, providing clearer directions for candidate refinement than scalar errors alone~\citep{li2026robust,pang2026deliberate}. However, their feedback mechanism depend on solver-specific search information, such as search trajectories memory, parent-selection strategy, and operator policy. Thus, feedback mechanisms are tightly coupled to the method's search process and remain an internal solver capability rather than reusable feedback. This coupling creates an undesirable trade-off: obtaining such enhanced feedback mechanism generally requires adopting the accompanying search framework, rather than retaining the distinctive mechanisms of existing solvers, such as LaSR's concept--equation co-evolution, SR-Scientist's autonomous tool-use loop, and PySR's efficient genetic-programming engine.

This motivates a different question: \emph{can richer diagnostic feedback be made reusable across diverse solvers, so that their shared feedback bottleneck can be addressed without sacrificing their distinctive search mechanisms?} Rather than introducing another new solver, we pursue a route orthogonal to solver design: constructing a diagnostic feedback mechanism that can be integrated with different SR solvers. Such a mechanism should be \textbf{search-independent}, neither depending on nor altering the solver's underlying search mechanism; \textbf{informative}, revealing where and how a candidate fails; and \textbf{solver-compatible}, allowing diagnosis-guided revisions to be readily returned to the solver.

\begin{figure}
    \centering
    \includegraphics[width=1\textwidth]{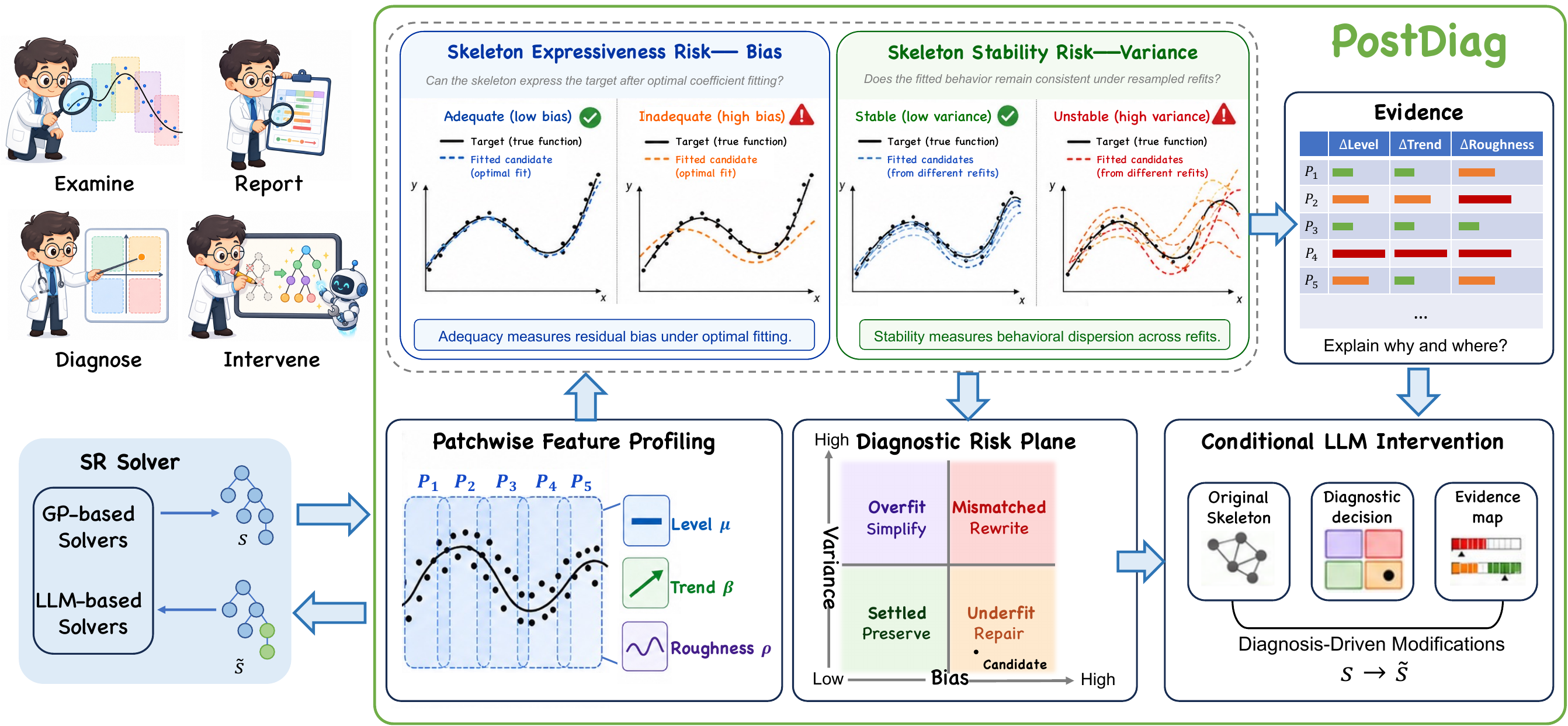}
    \caption{Overview of \textsc{PostDiag}, a search-independent post-diagnosis pipeline for symbolic regression. Given a candidate skeleton $s$ produced by LLM-based or GP-based SR solver, \textsc{PostDiag} first \emph{examines} its local behavior via patchwise feature profiling, then \emph{reports} the diagnostic report (evidence map), and \emph{diagnoses} the failure type on the diagnostic risk plane, and finally calls a conditional LLM \emph{intervention}, yielding a diagnosis-driven modification $\tilde{s}$ that is returned to the solver.}
    \vspace{-1.7em}
    \label{fig:intro}
\end{figure}

To instantiate this view, we propose \textbf{PostDiag}, a search-independent post-diagnostic module for SR. Unlike feedback mechanisms embedded in the search loop, PostDiag takes only candidate skeletons and observed data as input, without relying on solver-specific search mechanisms. It can therefore be integrated with solvers that employ different search mechanisms. For each candidate, PostDiag produces two complementary outputs: an evidence map that reveals where and how the candidate fails, and a risk localization that determines the appropriate intervention. PostDiag first introduces Patch-wise Feature Profiling (PFP), which represents local candidate behavior using structured fingerprints of \emph{level}, \emph{trend}, and \emph{roughness}. Rather than compressing diagnostic information into a single global error, discrepancies between these local fingerprints form an interpretable evidence map and support the estimation of \emph{Skeleton Expressiveness Risk} and \emph{Skeleton Stability Risk}. The former captures the structural mismatch remaining after coefficient fitting, while stability measures the candidate's sensitivity to data resampling. The Diagnostic Risk Plane (DRP) uses these two quantities to assign the candidate to one of four intervention regions: preserve, repair, simplify, or rewrite. A conditional LLM then performs the corresponding skeleton intervention, and the revised candidate is returned to the original solver through a lightweight adapter. We integrate the PostDiag with different SR solvers spanning LLM-based and GP-based methods, and evaluate them on the LSR-Synth dataset of LLM-SRBench. The results show that PostDiag improves candidate refinement and accelerates convergence without replacing the solvers' original search mechanisms. Ablation studies further validate the contribution of each diagnostic component.

Our contributions come in fourfold: (1) We introduce a solver-independent post-diagnostic perspective for SR that separates feedback design from search design, addressing a shared feedback bottleneck without replacing existing solvers. (2) We propose \textbf{PostDiag}, a reusable diagnostic module that takes only a candidate skeleton and observed data as input and returns evidence-supported diagnoses and candidate revisions through lightweight solver adapters. (3) We develop PFP to extract patch-wise diagnostic evidence and DRP to map skeleton adequacy and stability to targeted interventions. (4) We integrate the PostDiag with seven different LLM-based and GP-based SR solvers,  demonstrating broad compatibility and consistent convergence improvements.

\section{Related Work}
Symbolic regression (SR) spans diverse search and learning paradigms,
each contributing distinct inductive biases and capabilities
~\citep{chen2022symbolic,schmidt2009distilling,cranmer2020discovering,hu2026dass}.
Genetic-programming methods evolve expression trees through
population-based operators~\citep{gp,pawlak2018competent,
moraglio2012geometric,de2021interaction,gplearn}, with PySR providing
efficient multi-population search, complexity control, and constant
optimization~\citep{pysr}. Tree-search, reinforcement-learning, and
neural methods instead leverage planning, learned generation policies,
or pretrained representations~\citep{spl,dsr,transr,
mevznar2023efficient,holt2023deep,nsr,e2e,
shojaee2023transformer,meidani2023snip,gensr}. More recently,
LLM-based methods have incorporated scientific priors and reasoning:
LLM-SR combines program proposal with evolutionary experience
~\citep{llmsr}, LaSR co-evolves equations and a concept library
~\citep{lasr}, and many SR agent methods performs autonomous tool-assisted
optimization~\citep{xia2025sr,hu2026beyond,li2026learning}. Despite their diverse generation and
search mechanisms, these methods share a common feedback bottleneck:
candidate behavior is compressed into scalar fitness, which supports
ranking but provides little information about failure modes, structural
defects, or appropriate interventions.

Deliberate Evolution (DE)~\citep{pang2026deliberate} augments an MSE-driven evolutionary process with data and residual diagnostics, adaptive search operators, and trajectory-level memory, demonstrating that structured feedback can improve symbolic search beyond reliance on scalar errors alone. However, DE addresses this shared feedback bottleneck by introducing a new end-to-end solver, in which diagnosis is tightly coupled with its parent-selection strategy, operator policy, and trajectory memory. Adopting DE therefore entails replacing the original solving framework, making it difficult to retain solver-specific strengths such as LaSR’s concept–equation co-evolution, SR-Scientist’s agentic closed-loop optimization, or PySR’s efficient genetic-programming engine. Its limitation lies in addressing a common challenge across SR methods through a dedicated solver rather than a capability that existing solvers can readily share.

PostDiag takes a different technical route. Instead of replacing existing solvers, it provides a shared diagnostic interface decoupled from their internal search processes. Given only a candidate skeleton produced by the solver and the observation, PostDiag uses Patch-wise Feature Profiling to recover local behavioral information obscured by scalar errors and provide evidence for risk localization and candidate intervention. It can therefore enhance diagnostic feedback across heterogeneous SR solvers while preserving their original search mechanisms and distinctive strengths.

\section{A Search-Independent Post-Diagnosis for SR}

\subsection{Problem Setup}
Given a dataset $\mathcal{D}=\{(\bm{x}_i,y_i)\}_{i=1}^{N}$, where $\bm{x}_i\in\mathbb{R}^{d}$ and $y_i\in\mathbb{R}$, we represent a candidate expression as $f_s(\bm{x};\bm{\theta}_s)$. Here, $s$ denotes the symbolic skeleton and $\bm{\theta}_s$ is its numerical coefficients. PostDiag treats the skeleton $s$ as the object of diagnosis.

We randomly partition $\mathcal{D}$ into a disjoint fit set $\mathcal{D}_{\mathrm{fit}}$ and audit set $\mathcal{D}_{\mathrm{audit}}$. The fit set is used to estimate the numerical coefficients of the candidate skeleton, while the audit set is used to evaluate the fitted candidate and produce its diagnosis. This separation prevents coefficient fitting and candidate diagnosis from using the same observations.

Given a candidate skeleton $s$ and dataset $\mathcal{D}$, PostDiag outputs a structured diagnostic report $\mathcal{G}_s$ and a revised skeleton $\widetilde{s}$:
\[
\operatorname{PostDiag}(s,\mathcal{D})
\longrightarrow
\left(\mathcal{G}_s,\widetilde{s}\right).
\]
Here, $\mathcal{G}_s=(c_s,\mathcal{E}_s)$ contains the candidate's diagnostic coordinate $c_s$ on the diagnostic risk plane and supporting evidence $\mathcal{E}_s$, while $\widetilde{s}$ is the candidate skeleton revised according to the diagnosis.

PostDiag neither generates initial candidates nor prescribes how a solver conducts its search. It diagnoses candidate skeletons supplied by the solver and returns the diagnostic results and revised skeletons through a unified interface. The same diagnostic core can therefore be integrated with different SR solvers through lightweight adapters.

\subsection{Patch-wise Feature Profiling (PFP)}
To support expressiveness and stability estimation without compressing evaluation into a scalar error, PostDiag uses Patch-wise Feature Profiling to construct structured fingerprints on the observation. 

We first normalize each input dimension and denote the normalized coordinate of $\bm{x}_i$ by $\mathbf{x}_i$. 
We then construct $M$ local patches $\mathcal P=\{P_m\subseteq\mathcal D_{\mathrm{audit}}\}_{m=1}^{M}$. Such patches can be constructed in various ways. We adopt an anchor-$k$NN for its simplicity and robustness in multivariate settings. Specifically, we select spatially well-distributed anchors via farthest point sampling~\citep{gonzalez1985clustering} and construct each patch from the $k$ nearest neighbors~\citep{cover1967nearest} of an anchor $\hat{\mathbf{x}}_m$:
$$
P_m = \{(\mathbf{x}_i, y_i) \in \mathcal D_{\mathrm{audit}} \mid \mathbf{x}_i \text{ is among the } k \text{ nearest neighbors of } \hat{\mathbf{x}}_m \}.
$$
The patches may overlap, reducing sensitivity to neighborhood boundaries, while $M$ and $k$ are chosen to cover all audit samples.

Given a patch $P_m$, PFP summarizes the local behavior of $f$ using three complementary features.
\paragraph{Level.}
The level is the mean function value within the patch:
\begin{equation}
\mu_m(f)=\frac{1}{k}\sum_{i\in P_m}f(\bm{x}_i).
\end{equation}
Equivalently, $\mu_m(f)$ is the best constant approximation of $P_m$ in the least-squares sense, i.e., $\mu_m(f) = \arg\min_{c \in \mathbb R}\sum_{i \in P_m}(f({\bm x}_i)-c)^2$. Thus, $\mu_m(f)$ summarizes the zeroth-order behavior of $f$ on the patch $P_m$.

\paragraph{Trend.}
The trend captures the direction and magnitude in which $P_m$ varies across the normalized inputs. It is defined as the least-squares slope vector of $P_m$ on each direction:
$$
\bm{\beta}_m(f) =\operatorname*{arg\,min}_{\bm{\beta}\in\mathbb{R}^{d}}
\sum_{\mathbf{x}_i\in P_m}
\left[
f(\bm{x}_i)-\mu_m(f)
-\bm{\beta}^{\top}(\mathbf{x}_i-\bar{\mathbf{x}}_m)
\right]^2,
$$
where $\bar{\mathbf{x}}_m=\frac{1}{k}\sum_{\mathbf{x}_i\in P_m}\mathbf{x}_i$ is the normalized patch centroid. $\bm{\beta}_m(f)$ summarizes the first-order behavior of $f$ on the patch $P_m$. We retain ${\bm \beta}_m(f)$ as a $d$-dimensional vector to capture trends in multivariate symbolic regression. Estimating local slopes via least squares, rather than endpoint differences, further improves robustness to endpoint anomalies and observation noise.

\paragraph{Roughness.} Roughness captures the higher-order local behavior that cannot be explained by a constant level or a linear trend. For each $\bm{x}_i\in P_m$, we can define the local affine approximation use Level and Trend as
$$
\hat{f}_{m}(\bm x_i) = \mu_m(f) + {\bm \beta}_m(f)^{\top}({\mathbf{x}_i} - \bar{\mathbf{x}}_m).
$$
We therefore denote the residual between $f$ and $\hat{f}$ on $P_m$ as an additional signal:
$$
\rho_m(f)
  \;=\; \sqrt{\frac{1}{k}\sum_{i \in P_m}\bigl(f({\bm x}_i) - \hat{f}_m({\bm x}_i)\bigr)^{2}}.
$$
Accordingly, if $f$ is affine within $P_m$, its local behavior is fully explained by the Level and Trend terms, yielding $\rho_m(f)=0$. Local nonlinearity or curvature increases Roughness $\rho_m(f)$. 

The local fingerprint of $f$ on $P_m$ is
$$
\Phi_m(f)=\left(\mu_m(f),\bm{\beta}_m(f),\rho_m(f)\right),
$$
and its complete fingerprint over the $\mathcal{D}_{\rm audit}$ is
$$
\Phi(f)=\{\Phi_m(f)\}_{m=1}^{M}.
$$
By profiling different behaviors over the same patch partition, PFP provides a common basis for structured comparison. Each comparison is tied to a specific input-space patch and further decomposed into level, trend, and roughness features. It therefore reveals not only where behaviors differ, but also how they differ, while reducing reliance on individual pointwise residuals. In the next subsection, we use PFP as a comparison tool to assess skeleton adequacy and stability. Together, these comparisons yield the diagnostic coordinate $c_s$ and its supporting evidence map $\mathcal{E}_s$.

\subsection{Diagnostic Risk Plane}
PostDiag is designed to diagnose a candidate without depending on or altering the solver's search mechanism. It therefore cannot rely on solver-specific signals such as search trajectories, visitation statistics, population states, or operator histories, and must infer the appropriate intervention from only the candidate skeleton and observed data. To meet this constraint, we introduce the Diagnostic Risk Plane (DRP), which characterizes each skeleton using two behavior-based risks: \emph{Skeleton Expressiveness Risk} and \emph{Skeleton Stability Risk}. The former asks whether the skeleton can represent the observed behavior after coefficient fitting, whereas the latter asks whether the skeleton exhibits consistent behavior when its coefficients are re-estimated from different subset of the fitting data.

\textbf{Skeleton expressiveness risk.} We quantify this risk by a fingerprint-based bias score $\mathrm{Bias}(s)$. Specifically, let $\tilde{\bm{\theta}}_s$ denote the coefficients of skeleton $s$ fitted on $\mathcal{D}_{\mathrm{fit}}$, and let $\tilde{f}_s(\bm{x}) = f_s(\bm{x};\tilde{\bm{\theta}}_s)$. We evaluate $\tilde{f}_s$ on $\mathcal{D}_{\mathrm{audit}}$ and compare its fingerprint with that of the observed responses $\bm{y}_{\rm audit}$ over the same patches. For each patch $P_m$, we compute
$$
\Delta\mu_m=|\mu_m(\tilde{f}_s)-\mu_m(\bm{y}_{\mathrm{audit}})|, 
\Delta\bm{\beta}_m=\|\bm{\beta}_m(\tilde{f}_s)-\bm{\beta}_m(\bm{y}_{\mathrm{audit}})
\|_2, 
\Delta\rho_m=|\rho_m(\tilde{f}_s)-\rho_m(\bm{y}_{\mathrm{audit}})|.
$$
The local expressiveness risk is
$$
\mathrm{Bias}_m(s)=\omega_\mu\Delta\mu_m+\omega_\beta\Delta\bm{\beta}_m+\omega_\rho\Delta\rho_m,
$$
and the overall risk is
$$
\mathrm{Bias}(s)=\frac{1}{M}\sum_{m=1}^{M}\mathrm{Bias}_m(s).
$$
A low $\mathrm{Bias}(s)$ indicates that the fitted skeleton reproduces the observed behavior across patches, whereas a high value indicates insufficient expressiveness. The nonnegative weights $\omega_\mu$, $\omega_\beta$, and $\omega_\rho$ balance the three features. In practice, we assign a smaller weight to roughness because it is more sensitive to local fluctuations in finite patches.

\textbf{Skeleton stability risk.} Expressiveness alone cannot reveal whether the fitted behavior is reproducible or depends on the particular observations used for coefficient fitting. We characterize this dependence as skeleton stability risk. A stable skeleton should induce similar behavior when refitted on different resamples of the same observations, whereas a fragile skeleton relies on sample-specific coefficient configurations and therefore varies substantially across refits. We therefore quantify skeleton stability risk by the dispersion score $\mathrm{Var}(s)$ of the refitted function under different subset of the $\mathcal{D}_{\rm fit}$. Specifically, we generate $B$ fitting subsets $\{\mathcal{D}_b\}_{b=1}^{B}$ by bootstrap resampling~\citep{efron1992bootstrap} of the $\mathcal{D}_{fit}$ and refit the coefficients of the same skeleton $s$ on each subset. This yields \[f_s^{(b)}(\bm{x})=f_s(\bm{x};\bm{\theta}_s^{(b)}),\qquad b=1,\ldots,B\]. All refitted functions are evaluated on the same audit patches. For each patch $P_m$, we measure the dispersion of the fingerprints produced by the $\{f_s^{(b)}(\bm{x})\}_{b=1}^B$:
$$
\Sigma_{\mu,m}=\operatorname{std}_{b}[\mu_m(f_s^{(b)})],
\Sigma_{\rho,m}=\operatorname{std}_{b}[\rho_m(f_s^{(b)})], 
\Sigma_{\beta,m}=\frac{1}{B}\sum_{b=1}^{B}\|
\bm{\beta}_m(f_s^{(b)})-\overline{\bm{\beta}}_m\|_2,
$$
where $\overline{\bm{\beta}}_m=\frac{1}{B}\sum_{b=1}^{B}\bm{\beta}_m(f_s^{(b)})$. The local stability risks of patch $P_m$ are
$$
\mathrm{Var}_m(s)=\omega_\mu\Sigma_{\mu,m}+\omega_\beta\Sigma_{\beta,m}+\omega_\rho\Sigma_{\rho,m}, 
$$
and the overall stability risks of skeleton $s$ can be defined as
$$
\mathrm{Var}(s)=\frac{1}{M}\sum_{m=1}^{M}\mathrm{Var}_m(s).
$$
A larger $\mathrm{Var}(s)$ indicates greater variation in the fitted behavior across resamples and thus higher skeleton stability risk.

\paragraph{Risk localization and diagnostic evidence.} Each candidate $s$ is positioned in the Diagnostic Risk Plane (DRP) by the coordinate $c_s=\bigl(\mathrm{Bias}(f_s),\mathrm{Var}(f_s)\bigr)$. Meanwhile, the patch-level quantities are retained as the evidence map
$$
\mathcal{E}_s=\left\{\left(\Delta\mu_m,\Delta\bm{\beta}_m,\Delta\rho_m,\mathrm{Var}_m(s)
\right)\right\}_{m=1}^{M}.
$$
Thus, $c_s$ provides a compact risk localization for selecting an intervention, while $\mathcal{E}_s$ identifies the patches and behavioral features responsible for that diagnosis.

Using thresholds $\tau_B$ and $\tau_V$, the DRP distinguishes four diagnostic states, each corresponding to a different failure mode and intervention strategy:
\begin{itemize}
    \item \textbf{\textsc{Settled}}
    $(\mathrm{Bias}\leq\tau_B,\mathrm{Var}\leq\tau_V)$: the skeleton is sufficiently expressive and stable. The intervention is therefore preservation-oriented: the skeleton is either \textbf{preserved} or only minimally refined according to the evidence map.

    \item \textbf{\textsc{Underfit}}
    $(\mathrm{Bias}>\tau_B,\mathrm{Var}\leq\tau_V)$: the skeleton exhibits stable but systematic behavioral discrepancies, indicating missing or incorrect local structure. The candidate should be \textbf{repaired} by adding or replacing local symbolic components guided by the evidence map.

    \item \textbf{\textsc{Overfit}}
    $(\mathrm{Bias}\leq\tau_B,\mathrm{Var}>\tau_V)$: the skeleton matches the observations but is unstable under resampling, suggesting redundant or poorly identifiable structure. The candidate should be \textbf{simplified}.

    \item \textbf{\textsc{Mismatched}}
    $(\mathrm{Bias}>\tau_B,\mathrm{Var}>\tau_V)$: the skeleton is neither expressive nor stable, indicating a more fundamental structural mismatch. The candidate should be \textbf{rewritten}.
\end{itemize}
The diagnostic coordinate determines \emph{what} intervention is appropriate, while the evidence map indicates \emph{where and how} the candidate fails. Both rely only on the candidate skeleton and observed data,enabling targeted diagnosis independent of the solver's internal search process.

\subsection{Diagnosis-Guided Intervention and Solver Integration}
To reduce overhead, PostDiag first applies a low-cost NMSE-based gate to determine whether a candidate needs structured diagnosis. Candidates with either very low or very high errors are excluded from the PostDiag pipeline. Accurate candidates require no structural intervention, while poor candidates are left to the base solver's native search mechanism for elimination or regeneration. PostDiag therefore focuses on intermediate candidates, for which more informative feedback is needed to determine an appropriate intervention. NMSE serves only as a coarse screening signal, whereas PFP and DRP provide the evidence and diagnostic decision that guide the intervention.

For each candidate entering the diagnosis pipeline, PostDiag produces $\mathcal{G}_s=(c_s,\mathcal{E}_s)$. The risk region determined by $c_s$ specifies the intervention type, while $\mathcal{E}_s$ identifies the patches and behavioral features that motivate the intervention. An LLM-based post-editor receives the original skeleton $s$, $c_s$, and $\mathcal{E}_s$, and proposes a diagnosis-guided revision $\widetilde{s}$. The LLM acts only as a conditional skeleton editor and does not participate in the solver's candidate generation or search control. Notably, a candidate entering the diagnostic process may still be classified as \textsc{Settled}. Although its scalar error is not yet low enough for direct acceptance, the DRP indicates that its skeleton is sufficiently expressive and stable. PostDiag still provides the LLM editor with the original skeleton, the preservation-oriented diagnosis, and the evidence map $\mathcal{E}_s$, but instructs it to either retain the skeleton or make only minor edits supported by the evidence. 

We apply a lightweight validation before returning the proposed revision. Specifically, we refit the coefficients of $\widetilde{s}$ on $\mathcal{D}_{\mathrm{fit}}$ and evaluate its prediction error on $\mathcal{D}_{\mathrm{audit}}$. The revision is accepted if it reduces the error; otherwise, it is rejected and $\widetilde{s}=s$.

The revised skeleton can be integrated into different solvers via lightweight adapters. We consider two common solver families: GP-based and LLM-based solvers. For GP-based solvers, an accepted revision is inserted into the population as a repaired offspring. For LLM-based solvers, it replaces the original candidate, while $(s,\mathcal{G}_s,\widetilde{s})$ is appended to the prompt as a \emph{diagnostic revision record}. Additional adapter details are provided in Appendix~\ref{app:base-solver-integration}. PostDiag leaves candidate generation, selection, and exploration to the base solver, deriving diagnoses solely from the candidate skeleton and observed data without accessing solver-specific search states or histories.

\begin{table*}[t]
\color{black}
\caption{ID performance of base solvers and their PostDiag-enhanced variants on LSR-Synth. We report accuracy to tolerance, NMSE and runtime (the mean end-to-end wall-clock seconds per task run, including both SR base solver and PostDiag overhead) over all 129 tasks. All metrics are aggregated over three seeds. $A_\tau$ abbreviates $\mathrm{Acc}_\tau$.}
\label{tab:main}
\centering
\scriptsize
\setlength{\tabcolsep}{1.25pt}
\renewcommand{\arraystretch}{1.08}
\resizebox{\textwidth}{!}{%
\begin{tabular}{llcccccccccccccccc}
\toprule
\multirow{2}{*}{\textbf{Category}} & \multirow{2}{*}{\textbf{Method / setting}} & \multicolumn{4}{c}{\textbf{Overall}} & \multicolumn{3}{c}{\textbf{Phys}} & \multicolumn{3}{c}{\textbf{MatSci}} & \multicolumn{3}{c}{\textbf{Chem}} & \multicolumn{3}{c}{\textbf{Bio}} \\
\cmidrule(lr){3-6}\cmidrule(lr){7-9}\cmidrule(lr){10-12}\cmidrule(lr){13-15}\cmidrule(lr){16-18}
& & Time (s) & NMSE & $A_{.01}$ & $A_{.001}$ & NMSE & $A_{.01}$ & $A_{.001}$ & NMSE & $A_{.01}$ & $A_{.001}$ & NMSE & $A_{.01}$ & $A_{.001}$ & NMSE & $A_{.01}$ & $A_{.001}$ \\
\midrule
\multirow{10}{*}{LLM-based} & Direct LLM & 30.5 & 7.72E-02 & 2.6 & 0.8 & 7.32E-02 & 3.0 & 0.8 & 6.24E-02 & 2.7 & 1.3 & 3.50E-02 & 2.8 & 0.9 & 1.63E-01 & 1.4 & 0.0 \\
 & \textbf{Direct LLM+PostDiag} & 33.7 & \textbf{3.23E-02} & \textbf{4.4} & \textbf{2.3} & \textbf{5.22E-02} & \textbf{5.3} & \textbf{3.8} & \textbf{4.34E-03} & \textbf{4.0} & \textbf{2.7} & \textbf{8.76E-03} & \textbf{3.7} & 0.9 & \textbf{6.03E-02} & \textbf{4.2} & \textbf{1.4} \\
 & SGA & 320.0 & 1.39E-02 & 11.9 & 5.9 & 9.43E-03 & 8.3 & 6.1 & 8.21E-03 & 12.0 & 5.3 & 4.92E-03 & 19.4 & 7.4 & 4.15E-02 & 6.9 & 4.2 \\
 & \textbf{SGA+PostDiag} & 350.2 & \textbf{2.40E-03} & \textbf{18.6} & \textbf{9.8} & \textbf{2.66E-03} & \textbf{14.4} & \textbf{9.8} & \textbf{1.87E-04} & \textbf{25.3} & \textbf{6.7} & \textbf{8.65E-04} & \textbf{24.1} & \textbf{13.0} & \textbf{6.52E-03} & \textbf{11.1} & \textbf{8.3} \\
 & LLM-SR & 167.5 & 1.45E-03 & 22.0 & 8.8 & 3.66E-03 & 14.4 & 8.3 & 3.12E-05 & 37.3 & 13.3 & 1.75E-04 & 27.8 & 8.3 & 8.15E-04 & 11.1 & 5.6 \\
 & \textbf{LLM-SR+PostDiag} & 191.7 & \textbf{2.12E-04} & \textbf{31.0} & \textbf{15.2} & \textbf{3.48E-04} & \textbf{23.5} & \textbf{12.1} & \textbf{8.94E-06} & \textbf{60.0} & \textbf{30.7} & \textbf{5.84E-05} & \textbf{28.7} & \textbf{12.0} & \textbf{4.02E-04} & \textbf{18.1} & \textbf{9.7} \\
 & LaSR & 559.2 & 1.06E-02 & 13.4 & 7.5 & 1.73E-02 & 11.4 & 6.8 & 2.48E-03 & 14.7 & 6.7 & 6.90E-03 & 18.5 & 11.1 & 1.22E-02 & 8.3 & 4.2 \\
 & \textbf{LaSR+PostDiag} & 595.4 & \textbf{8.62E-04} & \textbf{21.4} & \textbf{12.7} & \textbf{1.52E-03} & \textbf{20.5} & \textbf{11.4} & \textbf{4.82E-05} & \textbf{29.3} & \textbf{14.7} & \textbf{6.45E-04} & \textbf{22.2} & \textbf{14.8} & \textbf{8.29E-04} & \textbf{13.9} & \textbf{9.7} \\
 & SR-Scientist & 648.0 & 1.12E-03 & 25.6 & 11.9 & 2.74E-03 & 17.4 & 9.8 & 4.47E-05 & 46.7 & 21.3 & 1.49E-04 & 27.8 & 11.1 & 7.26E-04 & 15.3 & 6.9 \\
 & \textbf{SR-Scientist+PostDiag} & 702.2 & \textbf{1.92E-04} & \textbf{35.7} & \textbf{20.4} & \textbf{3.11E-04} & \textbf{25.8} & \textbf{15.9} & \textbf{1.08E-05} & \textbf{62.7} & \textbf{37.3} & \textbf{6.47E-05} & \textbf{38.0} & \textbf{20.4} & \textbf{3.54E-04} & \textbf{22.2} & \textbf{11.1} \\
\midrule
\multirow{4}{*}{GP-based} & PySR & 8.8 & 2.67E-03 & 23.8 & 11.4 & 4.11E-03 & 21.2 & 12.1 & 1.70E-04 & 44.0 & 18.7 & 8.93E-04 & 20.4 & 10.2 & 5.29E-03 & 12.5 & 4.2 \\
 & \textbf{PySR+PostDiag} & 10.5 & \textbf{1.34E-03} & \textbf{28.7} & \textbf{13.2} & \textbf{1.82E-03} & \textbf{24.2} & \textbf{14.4} & 2.25E-04 & \textbf{50.7} & 17.3 & \textbf{5.29E-04} & \textbf{25.9} & \textbf{12.0} & \textbf{2.84E-03} & \textbf{18.1} & \textbf{8.3} \\
 & gplearn & 95.5 & 1.34E-01 & 0.0 & 0.0 & 8.51E-02 & 0.0 & 0.0 & 2.36E-02 & 0.0 & 0.0 & 6.56E-02 & 0.0 & 0.0 & 4.42E-01 & 0.0 & 0.0 \\
 & \textbf{gplearn+PostDiag} & 101.7 & \textbf{3.07E-02} & \textbf{1.0} & 0.0 & \textbf{4.29E-03} & \textbf{2.3} & 0.0 & \textbf{6.91E-03} & 0.0 & 0.0 & \textbf{4.29E-02} & \textbf{0.9} & 0.0 & \textbf{8.58E-02} & 0.0 & 0.0 \\
\bottomrule
\end{tabular}%
}
\end{table*}

\section{Experiment}
\subsection{Experimental Setup}\label{setup}
\textbf{Dataset.}
Because PostDiag and most primary baselines use LLMs, we evaluate them on the LSR-Synth subset of LLM-SRBench~\citep{llmsrbench}. Unlike LSR-Transform, which may remain vulnerable to pretraining contamination~\citep{xia2025sr}, LSR-Synth combines canonical domain terms with newly synthesized terms and filters tasks for solvability, novelty, and scientific plausibility. It contains 129 tasks across physics (44), materials science (25), chemistry (36), and biology (24), each with in-domain and out-of-domain test sets.

\textbf{Metrics.} We evaluate the discovered equations from 3 perspectives: numerical accuracy, accuracy-to-tolerance, and search efficiency. Numerical accuracy is measured by the normalized mean squared error (NMSE) between the predictions of the discovered equation $f$ and the ground truth:
$$
\mathrm{NMSE}(f)
=
\frac{
\sum_{i=1}^{N_{\mathrm{test}}} (\hat{y}_i - y_i)^2
}{
\sum_{i=1}^{N_{\mathrm{test}}} (y_i - \bar{y})^2
}.
$$
where $\hat{y}_i$ is the prediction target value and $y_i$ is the ground-truth value on $\bm x_i$. NMSE normalizes the error by the target variance, making it directly comparable across tasks with different output scales. We further report $\rm Acc_{0.01}$ following~\cite{e2e,llmsr,xia2025sr} to show the task-level solving accuracy at stringency levels:
$$
\mathrm{Acc}_{\tau}(f)
=
\mathds {1}
\left(
\max_{1 \le i \le N_{\mathrm{test}}}
\left|
\frac{\hat{y}_i - y_i}{y_i}
\right|
\le \tau
\right).
$$
We report $\mathrm{Acc}_{\tau}$ for $\tau\in\{.01,.001\}$~\cite{xia2025sr}; tables use $A_\tau$.
Search efficiency is measured by the average per-task search time (in seconds), which reflects the additional computational cost introduced by PostDiag. 

\textbf{Baselines.} PostDiag is a search-independent diagnostic module rather than a standalone solver; we therefore compare each base solver with its PostDiag-enhanced version. We evaluate five LLM-based solvers: LLM-SR~\citep{llmsr}, LaSR~\citep{lasr}, SR-Scientist~\citep{xia2025sr}, Direct
LLM~\citep{llmsrbench}, and SGA~\citep{ma2024llm}; and two GP-based solvers, PySR~\citep{pysr} and gplearn~\citep{gplearn}. Integration details are in Appendix~\ref{app:base-solver-integration}.

\textbf{Configuration of PostDiag.}  We describe the default configuration of PostDiag used in all experiments. For the data split, each training dataset $\mathcal{D}$ is randomly partitioned into a fitting set $\mathcal{D}_{\rm tr}$ and an audit set $\mathcal{D}_{\rm aud}$ with a $75\%/25\%$ ratio. On $\mathcal{D}_{\rm aud}$, PostDiag in the experiments of LSR-Synth constructs $M = 64$ overlapping anchor-$k$NN patches using farthest-point anchors, where $k = 64$. We suggest that the number of anchors and the neighborhood size should be scaled to the audit-set size to guarantee that every audit sample belongs to at least one patch. For fingerprint discrepancies, the three feature dimensions are combined with weights $(\omega_\mu, \omega_\beta, \omega_\rho) = (0.4, 0.4, 0.2)$; the smaller weight on roughness reflects its higher sensitivity to observation noise on small patches. Skeleton stability is estimated from $B=3$ bootstrap refits, each of size $\lfloor \rho\,|\mathcal{D}_{\rm fit}| \rfloor$ points drawn with replacement, where $\rho = 0.7$. The Diagnostic Risk Plane thresholds are setted as $\tau_B = 0.25$ and $\tau_V = 0.15$ based on preliminary experiments on a small held-out subset of tasks. We use these thresholds transfers across all four benchmark domains without per-domain re-tuning.

\textbf{Search budget.}
To evaluate the effectiveness of PostDiag under limited search resources, we adopt low-budget configurations in the main experiments. For each task, LLM-SR is limited to 100 LLM calls; LaSR runs for 20 outer iterations; SR-Scientist runs for 10 outer iterations, each with at most 25 interaction turns; PySR runs for 65 iterations; and gplearn evolves for 100 generations. Direct
LLM generates 5 candidates in Low and 10 in High from a single prompt batch; SGA runs 5 rounds (80 generated candidates) in Low and 20 rounds (320 generated candidates) in High. For each pair of a base solver and its PostDiag-augmented variant, we keep the base search budget, internal hyperparameters, LLM backbone, and random seeds identical. The diagnostic and structural intervention costs introduced by PostDiag are not deducted from the base solver's search budget, but are included in the reported end-to-end runtime. Measured results are aggregated over three random seeds. Table~\ref{tab:main} uses Qwen3.5-27B for all LLM-based solvers and for PostDiag interventions on GP solvers; Appendix~\ref{app:additional-results} reports the Qwen3.5-122B-A10B comparison, symbolic accuracy, budget efficiency, and additional backbones, while Appendix~\ref{app:ood} reports OOD results.

\subsection{Results of Different SR Solvers Integrated with PostDiag}
Table~\ref{tab:main} compares seven SR solvers with and without PostDiag on LSR-Synth. Across LLM-based solvers, model scales, and domains, PostDiag consistently reduces NMSE and improves $\mathrm{Acc}_{0.01}$, indicating that stronger LLM backbones do not eliminate the need for structured diagnostic feedback. The improvements also extend to GP-based solvers, although it cannot fully compensate for the gplearn's limited search capacity. These gains incur additional diagnostic cost from PostDiag, but require no modification to the underlying search mechanisms.

We evaluate all solvers under reduced search budgets to test whether PostDiag improves search efficiency. Larger budgets can mask differences in discovery speed as solvers approach convergence. Despite using far fewer search steps, PostDiag-enhanced solvers match or exceed results reported with much larger budgets. For example, in SR-Scientist~\citep{xia2025sr}, the authors report \(\mathrm{Acc}_{0.01}\) of 30.49 for LLM-SR and 29.46 for their training-free variant with Qwen3-32B, using 1,000 LLM calls and 40 outer iterations. With Qwen3.5-27B, LLM-SR+PostDiag and SR-Scientist+PostDiag reach 31.0 and 35.7 using only 100 calls and 10 iterations, respectively, indicating improved search efficiency rather than gains from larger budgets.

\subsection{Diagnostic Fidelity of the PostDiag}
\label{sec:drp-fidelity}
To assess whether PostDiag correctly identifies skeleton risks, we manually construct four labeled skeleton–data pair types. Settled pairs use the correct skeleton with varying Gaussian noise; Underfit pairs preserve the main structure and remove or replace one true term; Overfit pairs retain the correct skeleton and add redundant or unstable terms; and Mismatched pairs pair the data with an unrelated skeleton. Without access to these labels, PostDiag estimates Skeleton Bias and Variance via PFP and maps each candidate onto the DRP.

As shown in Figure~\ref{fig:diagnostic-fidelity-drp}, the four pair types largely occupy their expected DRP regions. We observe a moderate overlap between underfit and mismatch near the stability boundary. This is expected, since both cases involve structural bias. When this bias is severe, coefficient refitting on resampled subsets may become unstable, making the boundary between an incomplete skeleton and a mismatched skeleton less sharp. Nevertheless, the clear quadrant structure shows that PFP and DRP provide a meaningful risk localization for different skeleton failure modes, yielding feedback with more explicit diagnostic information.

\subsection{Ablation Study}
\label{sec:intervention-ablation}
\begin{figure}[t]
    \centering
    \begin{subfigure}[t]{0.48\linewidth}
        \centering
        \includegraphics[width=\linewidth]{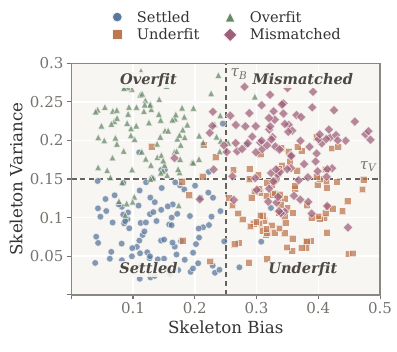}
        \vspace{-2em}
        \caption{}
        \label{fig:diagnostic-fidelity-drp}
    \end{subfigure}\hfill
    \begin{subfigure}[t]{0.48\linewidth}
        \centering
        \includegraphics[width=\linewidth]{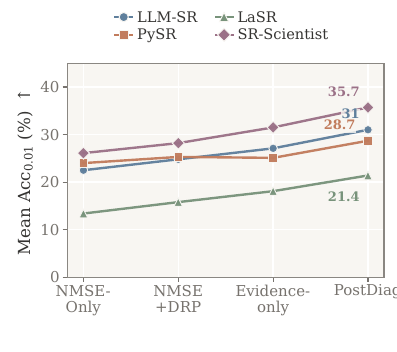}
        \vspace{-2em}
        \caption{}
        \label{fig:intervention-ablation}
    \end{subfigure}
    \caption{
    (a) Diagnostic fidelity of PostDiag. PFP-derived bias and variance scores map synthesis skeletons to distinct DRP regions, separating settled, underfit, overfit, and mismatched cases. (b) Component ablation across four base SR solvers. Each point reports pooled \(\mathrm{Acc}_{0.01}\) over 129 LSR-Synth tasks and three seeds. Full PostDiag performs best by combining DRP localization with evidence-guided intervention.
    }
    \label{fig:diagnosis-and-ablation}
\end{figure}

To assess the contribution of different diagnostic signals, we compare three ablated variants with full PostDiag. ``NMSE-only'' provides only scalar candidate error to the LLM intervention module. ``NMSE+DRP'' infers skeleton adequacy and stability from NMSE-based signals and places the candidate on the DRP. ``Evidence-only'' supplies the PFP-based evidence map without DRP localization. ``PostDiag'' combines the evidence map, DRP localization, and explicit instructions to preserve, repair, simplify, or rewrite the skeleton. We evaluate all variants on LLM-SR, LaSR, SR-Scientist, and PySR. Results in Figure~\ref{fig:intervention-ablation} are averaged across the four LSR-Synth domains. Qwen-27B serves as both the backbone for all LLM-based solvers and the intervention model for all four solvers.

As shown in Figure~\ref{fig:intervention-ablation}, NMSE-only yields little improvement because the base solvers already rely on scalar fitness, so passing NMSE to the intervention model provides little additional information. NMSE+DRP improves the results, but its gains are limited. Estimating skeleton risk from a scalar NMSE is too coarse to reliably distinguish different failure modes, which can lead to inaccurate DRP localization. Evidence-only produces much larger improvements for LaSR and SR-Scientist, suggesting that these solvers benefit more from detailed information about where and how a candidate fails than from a coarse risk label. For PySR, NMSE+DRP is slightly more effective than Evidence-only, indicating that an explicit intervention operator is useful for guiding its population-based search. Full PostDiag performs best across all solvers, confirming that PFP, the evidence map, and DRP are complementary: PFP localizes risk, the evidence map reveals local defects, and DRP selects the appropriate structural intervention.

\section{Conclusion and Limitations}
This work addresses a feedback bottleneck shared by diverse SR solvers: scalar fitness can rank candidates but offers little guidance for structural refinement. Rather than developing a new solver, PostDiag separates feedback design from search design. Using only a candidate skeleton and observed data, it converts patch-wise behavior into expressiveness and stability risks, localized evidence, and targeted interventions. Experiments across LLM-based and GP-based solvers show that PostDiag improves accuracy and accelerates convergence without accessing or redesigning their internal search processes. These results establish post-diagnosis as an orthogonal way to strengthen existing SR solvers while preserving their distinctive search mechanisms.

Despite its search-independent diagnostic core, our current implementation is evaluated only on LLM-based and GP-based solvers, which provide natural interfaces for accepting revised candidates. Solvers without such interfaces, such as reinforcement-learning-based methods~\citep{dsr}, may require more substantial adaptation. Future work will extend the diagnostic interface to broader search paradigms and improve the efficiency of diagnosis and intervention.

\section*{The Use of Large Language Models (LLMs)}
Beyond the use of LLMs in the method described in this paper, LLMs served only as auxiliary tools for generating some of the cartoon characters in Figure~\ref{fig:intro} and for polishing and refining the manuscript’s language. Their contributions to research conception, experimental design, or manuscript preparation do not warrant authorship or contributor status. All substantive research ideas, methods, and results presented in this work were independently developed or produced by the authors.

\bibliography{iclr2027_conference}

@inproceedings{gensr,
  title={GenSR: Symbolic regression based on equation generative space},
  author={Li, Qian and Hu, Yuxiao and Liu, Juncheng and Chen, Yuntian},
  booktitle={The Fourteenth International Conference on Learning Representations}
}

@article{xu2023retention,
  title={Retention time prediction for chromatographic enantioseparation by quantile geometry-enhanced graph neural network},
  author={Xu, Hao and Lin, Jinglong and Zhang, Dongxiao and Mo, Fanyang},
  journal={Nature Communications},
  volume={14},
  number={1},
  pages={3095},
  year={2023},
  publisher={Nature Publishing Group UK London}
}

@article{chen2022symbolic,
  title={Symbolic genetic algorithm for discovering open-form partial differential equations (SGA-PDE)},
  author={Chen, Yuntian and Luo, Yingtao and Liu, Qiang and Xu, Hao and Zhang, Dongxiao},
  journal={Physical Review Research},
  volume={4},
  number={2},
  pages={023174},
  year={2022},
  publisher={APS}
}

@article{lasr,
  title={Symbolic regression with a learned concept library},
  author={Grayeli, Arya and Sehgal, Atharva and Costilla Reyes, Omar and Cranmer, Miles and Chaudhuri, Swarat},
  journal={Advances in Neural Information Processing Systems},
  volume={37},
  pages={44678--44709},
  year={2024}
}

@inproceedings{llmsr,
  title={LLM-SR: Scientific Equation Discovery via Programming with Large Language Models},
  author={Shojaee, Parshin and Meidani, Kazem and Gupta, Shashank and Farimani, Amir Barati and Reddy, Chandan K},
  booktitle={The Thirteenth International Conference on Learning Representations},
  year={2025}
}

@article{pysr,
  title={Interpretable machine learning for science with PySR and SymbolicRegression. jl},
  author={Cranmer, Miles},
  journal={arXiv preprint arXiv:2305.01582},
  year={2023}
}

@inproceedings{transr,
  title={Transformer-based model for symbolic regression via joint supervised learning},
  author={Li, Wenqiang and Li, Weijun and Sun, Linjun and Wu, Min and Yu, Lina and Liu, Jingyi and Li, Yanjie and Tian, Songsong},
  booktitle={The Eleventh International Conference on Learning Representations},
  year={2022}
}

@inproceedings{spl,
  title={Symbolic Physics Learner: Discovering governing equations via Monte Carlo tree search},
  author={Sun, Fangzheng and Liu, Yang and Wang, Jian-Xun and Sun, Hao},
  booktitle={The Eleventh International Conference on Learning Representations},
  year={2023}
}

@inproceedings{dsr,
  title={Deep symbolic regression: Recovering mathematical expressions from data via risk-seeking policy gradients},
  author={Petersen, Brenden K and Larma, Mikel Landajuela and Mundhenk, Terrell N and Santiago, Claudio Prata and Kim, Soo Kyung and Kim, Joanne Taery},
  booktitle={International Conference on Learning Representations},
  year={2021}
}

@inproceedings{nsr,
  title={Neural symbolic regression that scales},
  author={Biggio, Luca and Bendinelli, Tommaso and Neitz, Alexander and Lucchi, Aurelien and Parascandolo, Giambattista},
  booktitle={International Conference on Machine Learning},
  pages={936--945},
  year={2021},
  organization={Pmlr}
}

@article{e2e,
  title={End-to-end symbolic regression with transformers},
  author={Kamienny, Pierre-Alexandre and d'Ascoli, St{\'e}phane and Lample, Guillaume and Charton, Fran{\c{c}}ois},
  journal={Advances in Neural Information Processing Systems},
  volume={35},
  pages={10269--10281},
  year={2022}
}

@article{gplearn,
  title={gplearn: Genetic programming in python, with a scikit-learn inspired api, 2024},
  author={Stephens, Trevor},
  journal={URL https://github. com/trevorstephens/gplearn. Accessed},
  pages={05--22},
  year={2024}
}

@inproceedings{zhang2025rag,
  title={RAG-SR: Retrieval-augmented generation for neural symbolic regression},
  author={Zhang, Hengzhe and Chen, Qi and Banzhaf, Wolfgang and Zhang, Mengjie and others},
  booktitle={The Thirteenth International Conference on Learning Representations},
  year={2025}
}

@article{shojaee2023transformer,
  title={Transformer-based planning for symbolic regression},
  author={Shojaee, Parshin and Meidani, Kazem and Barati Farimani, Amir and Reddy, Chandan},
  journal={Advances in Neural Information Processing Systems},
  volume={36},
  pages={45907--45919},
  year={2023}
}

@article{meidani2023snip,
  title={Snip: Bridging mathematical symbolic and numeric realms with unified pre-training},
  author={Meidani, Kazem and Shojaee, Parshin and Reddy, Chandan K and Farimani, Amir Barati},
  journal={arXiv preprint arXiv:2310.02227},
  year={2023}
}

@article{xia2025sr,
  title={Sr-scientist: Scientific equation discovery with agentic ai},
  author={Xia, Shijie and Sun, Yuhan and Liu, Pengfei},
  journal={arXiv preprint arXiv:2510.11661},
  year={2025}
}

@article{llmsrbench,
  title={Llm-srbench: A new benchmark for scientific equation discovery with large language models},
  author={Shojaee, Parshin and Nguyen, Ngoc-Hieu and Meidani, Kazem and Farimani, Amir Barati and Doan, Khoa D and Reddy, Chandan K},
  journal={arXiv preprint arXiv:2504.10415},
  year={2025}
}

@article{yu2025discover,
  title={Discover network dynamics with neural symbolic regression},
  author={Yu, Zihan and Ding, Jingtao and Li, Yong},
  journal={Nature Computational Science},
  pages={1--13},
  year={2025},
  publisher={Nature Publishing Group US New York}
}

@article{guo2025sr,
  title={SR-LLM: An incremental symbolic regression framework driven by LLM-based retrieval-augmented generation},
  author={Guo, Zelin and Wang, Siqi and Tian, Yonglin and Yang, Jing and Yu, Hui and Na, Xiaoxiang and Kov{\'a}cs, Levente and Li, Li and Ioannou, Petros A and Wang, Fei-Yue},
  journal={Proceedings of the National Academy of Sciences},
  volume={122},
  number={52},
  pages={e2516995122},
  year={2025},
  publisher={National Academy of Sciences}
}

@article{ma2024llm,
  title={Llm and simulation as bilevel optimizers: A new paradigm to advance physical scientific discovery},
  author={Ma, Pingchuan and Wang, Tsun-Hsuan and Guo, Minghao and Sun, Zhiqing and Tenenbaum, Joshua B and Rus, Daniela and Gan, Chuang and Matusik, Wojciech},
  journal={arXiv preprint arXiv:2405.09783},
  year={2024}
}

@article{gp,
  title={Genetic programming as a means for programming computers by natural selection},
  author={Koza, John R},
  journal={Statistics and computing},
  volume={4},
  number={2},
  pages={87--112},
  year={1994},
  publisher={Springer}
}

@misc{qwen35,
  title        = {{Qwen3.5}: Towards Native Multimodal Agents},
  author       = {{Qwen Team}},
  month        = feb,
  year         = {2026},
  howpublished = {\url{https://qwen.ai/blog?id=qwen3.5}},
  note         = {Accessed: 2026-07-22}
}

@article{schmidt2009distilling,
  title={Distilling free-form natural laws from experimental data},
  author={Schmidt, Michael and Lipson, Hod},
  journal={science},
  volume={324},
  number={5923},
  pages={81--85},
  year={2009},
  publisher={American Association for the Advancement of Science}
}

@article{cranmer2020discovering,
  title={Discovering symbolic models from deep learning with inductive biases},
  author={Cranmer, Miles and Sanchez Gonzalez, Alvaro and Battaglia, Peter and Xu, Rui and Cranmer, Kyle and Spergel, David and Ho, Shirley},
  journal={Advances in neural information processing systems},
  volume={33},
  pages={17429--17442},
  year={2020}
}

@article{wang2023scientific,
  title={Scientific discovery in the age of artificial intelligence},
  author={Wang, Hanchen and Fu, Tianfan and Du, Yuanqi and Gao, Wenhao and Huang, Kexin and Liu, Ziming and Chandak, Payal and Liu, Shengchao and Van Katwyk, Peter and Deac, Andreea and others},
  journal={Nature},
  volume={620},
  number={7972},
  pages={47--60},
  year={2023},
  publisher={Nature Publishing Group UK London}
}

@article{zhang2025artificial,
  title={Artificial intelligence for science in quantum, atomistic, and continuum systems},
  author={Zhang, Xuan and Wang, Limei and Helwig, Jacob and Luo, Youzhi and Fu, Cong and Xie, Yaochen and Liu, Meng and Lin, Yuchao and Xu, Zhao and Yan, Keqiang and others},
  journal={Foundations and Trends{\textregistered} in Machine Learning},
  volume={18},
  number={4},
  pages={385--849},
  year={2025},
  publisher={Emerald Publishing Limited}
}

@article{karniadakis2021physics,
  title={Physics-informed machine learning},
  author={Karniadakis, George Em and Kevrekidis, Ioannis G and Lu, Lu and Perdikaris, Paris and Wang, Sifan and Yang, Liu},
  journal={Nature Reviews Physics},
  volume={3},
  number={6},
  pages={422--440},
  year={2021},
  publisher={Nature Publishing Group UK London}
}

@article{krenn2022scientific,
  title={On scientific understanding with artificial intelligence},
  author={Krenn, Mario and Pollice, Robert and Guo, Si Yue and Aldeghi, Matteo and Cervera-Lierta, Alba and Friederich, Pascal and dos Passos Gomes, Gabriel and H{\"a}se, Florian and Jinich, Adrian and Nigam, AkshatKumar and others},
  journal={Nature Reviews Physics},
  volume={4},
  number={12},
  pages={761--769},
  year={2022},
  publisher={Nature Publishing Group UK London}
}

@article{udrescu2020ai,
  title={AI Feynman: A physics-inspired method for symbolic regression},
  author={Udrescu, Silviu-Marian and Tegmark, Max},
  journal={Science advances},
  volume={6},
  number={16},
  pages={eaay2631},
  year={2020},
  publisher={American Association for the Advancement of Science}
}

@article{udrescu2020ai2,
  title={AI Feynman 2.0: Pareto-optimal symbolic regression exploiting graph modularity},
  author={Udrescu, Silviu-Marian and Tan, Andrew and Feng, Jiahai and Neto, Orisvaldo and Wu, Tailin and Tegmark, Max},
  journal={Advances in Neural Information Processing Systems},
  volume={33},
  pages={4860--4871},
  year={2020}
}

@incollection{efron1992bootstrap,
  title={Bootstrap methods: another look at the jackknife},
  author={Efron, Bradley},
  booktitle={Breakthroughs in statistics: Methodology and distribution},
  pages={569--593},
  year={1992},
  publisher={Springer}
}

@article{openai2025gptoss120bgptoss20bmodel,
  title={gpt-oss-120b \& gpt-oss-20b model card},
  author={Agarwal, Sandhini and Ahmad, Lama and Ai, Jason and Altman, Sam and Applebaum, Andy and Arbus, Edwin and Arora, Rahul K and Bai, Yu and Baker, Bowen and Bao, Haiming and others},
  journal={arXiv preprint arXiv:2508.10925},
  year={2025}
}

@article{gonzalez1985clustering,
  title={Clustering to minimize the maximum intercluster distance},
  author={Gonzalez, Teofilo F},
  journal={Theoretical computer science},
  volume={38},
  pages={293--306},
  year={1985},
  publisher={Elsevier}
}

@article{cover1967nearest,
  title={Nearest neighbor pattern classification},
  author={Cover, Thomas and Hart, Peter},
  journal={IEEE transactions on information theory},
  volume={13},
  number={1},
  pages={21--27},
  year={1967},
  publisher={IEEE}
}

@article{mevznar2023efficient,
  title={Efficient generator of mathematical expressions for symbolic regression.},
  author={Me{\v{z}}nar, Sebastian and D{\v{z}}eroski, Sa{\v{s}}o and Todorovski, Ljup{\v{c}}o},
  journal={Machine Learning},
  volume={112},
  number={11},
  year={2023}
}

@article{holt2023deep,
  title={Deep generative symbolic regression},
  author={Holt, Samuel and Qian, Zhaozhi and van der Schaar, Mihaela},
  journal={arXiv preprint arXiv:2401.00282},
  year={2023}
}

@inproceedings{moraglio2012geometric,
  title={Geometric semantic genetic programming},
  author={Moraglio, Alberto and Krawiec, Krzysztof and Johnson, Colin G},
  booktitle={International Conference on Parallel Problem Solving from Nature},
  pages={21--31},
  year={2012},
  organization={Springer}
}

@article{pawlak2018competent,
  title={Competent geometric semantic genetic programming for symbolic regression and boolean function synthesis},
  author={Pawlak, Tomasz P and Krawiec, Krzysztof},
  journal={Evolutionary computation},
  volume={26},
  number={2},
  pages={177--212},
  year={2018},
  publisher={MIT Press}
}

@article{de2021interaction,
  title={Interaction--transformation evolutionary algorithm for symbolic regression},
  author={de Franca, Fabricio Olivetti and Aldeia, Guilherme Seidyo Imai},
  journal={Evolutionary computation},
  volume={29},
  number={3},
  pages={367--390},
  year={2021},
  publisher={MIT Press One Rogers Street, Cambridge, MA 02142-1209, USA journals-info~…}
}

@article{pang2026deliberate,
  title={Deliberate Evolution: Agentic Reasoning for Sample-Efficient Symbolic Regression with LLMs},
  author={Pang, Xinyu and Zhou, Zhanke and Li, Xuan and Lv, Fangrui and Wei, Shanshan and Cui, Sen and Han, Bo and Zhang, Changshui},
  journal={arXiv preprint arXiv:2606.04360},
  year={2026}
}

@inproceedings{xiangmod,
  title={MOD-SR: Unifying Multimodal Learning and Direct Optimization with Gradient-Guided Diffusion Model for Symbolic Regression},
  author={Xiang, Chuyang and Wei, Yichen and Yan, Junchi},
  booktitle={Forty-third International Conference on Machine Learning}
}

@inproceedings{li2026robust,
  title={Robust Equation Structure Learning with Adaptive Refinement},
  author={Li, Yunlun and Pan, Sinno Jialin},
  booktitle={International Conference on Learning Representations},
  volume={2026},
  pages={70029--70053},
  year={2026}
}

@article{hu2026beyond,
  title={Beyond Equation Search: Conjecture-Guided Discovery with SCOPE and ClosureBench},
  author={Hu, Yuxiao and Li, Qian and Wang, Ding and Chen, Siheng and Zhang, Linfeng and Yan, Jinyue and Chen, Yuntian},
  journal={arXiv preprint arXiv:2610.18938},
  year={2026}
}

@article{hu2026dass,
  title={DASS: A Solver-Adaptable Dynamic Auxiliary Search Strategy for Symbolic Regression},
  author={Hu, Yuxiao and Li, Qian and Wang, Ding and Zheng, Yekun and Yan, Jinyue and Chen, Yuntian},
  journal={arXiv preprint arXiv:2610.16812},
  year={2026}
}

@article{li2026learning,
  title={Learning to Design SR Solvers: Harness Optimization for Symbolic Regression},
  author={Li, Qian and Hu, Yuxiao and Duan, Qihao and Yan, Jinyue and Chen, Yuntian},
  journal={arXiv preprint arXiv:2610.16966},
  year={2026}
}

@inproceedings{hu2025context,
  title={Context-alignment: Activating and enhancing LLMs capabilities in time series},
  author={Hu, Yuxiao and Li, Qian and Zhang, Dongxiao and Yan, Jinyue and Chen, Yuntian},
  booktitle={International Conference on Learning Representations},
  volume={2025},
  pages={90696--90722},
  year={2025}
}

@article{li2023adaptive,
  title={Adaptive Perona--Malik model based on dynamical threshold for image multi--noise removal with details preservation},
  author={Li, Qian and Hu, Yuxiao and Cao, Yang},
  journal={Computers \& Mathematics with Applications},
  volume={137},
  pages={28--43},
  year={2023},
  publisher={Elsevier}
}

@article{hu2024domain,
  title={Domain knowledge-enhanced multi-spatial multi-temporal PM2. 5 forecasting with integrated monitoring and reanalysis data},
  author={Hu, Yuxiao and Li, Qian and Shi, Xiaodan and Yan, Jinyue and Chen, Yuntian},
  journal={Environment international},
  volume={192},
  pages={108997},
  year={2024},
  publisher={Elsevier}
}

@inproceedings{li2023discrete,
  title={Discrete point-wise attack is not enough: Generalized manifold adversarial attack for face recognition},
  author={Li, Qian and Hu, Yuxiao and Liu, Ye and Zhang, Dongxiao and Jin, Xin and Chen, Yuntian},
  booktitle={2023 IEEE/CVF Conference on Computer Vision and Pattern Recognition (CVPR)},
  pages={20575--20584},
  year={2023},
  organization={IEEE}
}

@inproceedings{li2024focus,
  title={Focus on hiders: Exploring hidden threats for enhancing adversarial training},
  author={Li, Qian and Hu, Yuxiao and Dong, Yinpeng and Zhang, Dongxiao and Chen, Yuntian},
  booktitle={2024 IEEE/CVF Conference on Computer Vision and Pattern Recognition (CVPR)},
  pages={24442--24451},
  year={2024},
  organization={IEEE}
}

@article{xu2025exploring,
  title={Exploring terrain-precipitation relationships with interpretable AI for advancing future climate projections},
  author={Xu, Hao and Chen, Yuntian and Zeng, Zhenzhong and Li, Nina and Li, Jian and Zhang, Dongxiao},
  journal={Nexus},
  volume={2},
  number={1},
  year={2025},
  publisher={Elsevier}
}

@article{xu2024worth,
  title={Worth of prior knowledge for enhancing deep learning},
  author={Xu, Hao and Chen, Yuntian and Zhang, Dongxiao},
  journal={Nexus},
  volume={1},
  number={1},
  year={2024},
  publisher={Elsevier}
}
\bibliographystyle{iclr2027_conference}
\newpage
\appendix

% \section{Reproducibility Statement}
% For the reproducibility of our results, we have provided a detailed description of our methods and experimental setups in Sec.~\ref{setup} and Appendix~\ref{app:base-solver-integration}, ~\ref{app:base-solvers},~\ref{app:compute}. We also confirmed the robustness of our results through the extensive experiments. In addition, to further facilitate the reproduction, we will release our codes and the checkpoints for the trained models.
\section{Dataset Details}
\label{app:dataset}

\subsection{Task Composition}

We conduct all experiments on LSR-Synth, the synthetic subset of
LLM-SRBench~\citep{llmsrbench}. Following the evaluation setting used by
SR-Scientist~\citep{xia2025sr}, the benchmark contains 129 symbolic
regression tasks from four scientific domains: 44 in physics, 25 in
material science, 36 in chemistry, and 24 in biology.

Representative tasks include damped oscillators, stress--strain
relations, chemical kinetics, and population dynamics. Each task provides numerical
observations together with the variable names and their scientific
meanings. Its ground-truth expression combines established
domain-specific terms with novel synthetic terms. Consequently, a solver
cannot rely solely on memorized canonical equations and must instead use
the observations to search for previously unseen structures. Compared
with LSR-Transform, which may be more susceptible to memorization from
pretraining, LSR-Synth is therefore better suited to evaluating
data-driven equation discovery.

\subsection{Data Generation and Splits}

The construction of LSR-Synth consists of scientific-problem selection,
known- and synthetic-term generation, parameter instantiation, numerical
solvability and novelty checks, data generation, and expert validation.
For each scientific problem, candidate known and synthetic terms are
first generated and combined into complete expressions, whose
coefficients are instantiated within scientifically meaningful ranges.
For the dynamical systems in chemistry, biology, and physics, data are
generated through numerical integration using SciPy's
\texttt{solve\_ivp} function with the RK45 method. The static
stress--strain relationships in material science are evaluated directly
over predefined input ranges. Candidates that cannot be solved reliably,
produce divergent or degenerate behavior, or are excessively similar to
existing tasks are filtered out. The retained expressions and
visualizations of their generated data are then cross-checked by domain
experts.

Each task contains 5,000 observations. Chemistry, biology, and physics
tasks are sampled uniformly over the time interval $t\in[0,60]$, whereas
material-science tasks cover strain $\epsilon\in[0,0.6]$ and temperature
$T\in[273,573]\,\mathrm{K}$. For time-dependent tasks, the 500
observations with the largest time values form the out-of-distribution
(OOD) test set. For material-science tasks, the 500 observations with the
largest temperature values form the OOD test set. Of the remaining 4,500
observations, 500 are randomly selected as the in-distribution (ID) test
set, and the other 4,000 observations constitute the training set
provided to the solver.

\subsection{Data-Usage Protocol for PostDiag}

All diagnosis and intervention operations performed by PostDiag are
restricted to the training split and are strictly isolated from both
test sets. Under the default configuration described in the main text,
the training data for each task are further divided into a fitting set
$\mathcal{D}_{\rm tr}$ and an audit set $\mathcal{D}_{\rm aud}$ using a
$75\%/25\%$ split. Candidate coefficient fitting and bootstrap refitting
use only $\mathcal{D}_{\rm tr}$, whereas the PFP fingerprints, skeleton
adequacy and stability scores, and evidence maps are computed on
$\mathcal{D}_{\rm aud}$. The acceptance check for an intervened candidate
also uses only the training split.

The ID and OOD test sets are used solely to evaluate the final expression
after search terminates. They are never used for candidate generation,
diagnosis, structural intervention, rewrite acceptance, or candidate
selection. Although an LLM-based base solver may consume the task and
variable descriptions supplied by the benchmark, PostDiag itself receives
only a candidate skeleton and the numerical observations from the training
split.

\section{Base Solver Configurations}
\label{app:base-solvers}
Recent progress has led to increasing interest in symbolic regression (SR)~\citep{chen2022symbolic,xu2023retention,meidani2023snip,zhang2025rag,li2023discrete,e2e,li2024focus,xia2025sr,llmsr}. Unlike conventional predictive models, SR seeks to recover explicit mathematical relationships from observations, making the resulting equations directly interpretable and scientifically testable across domains such as physics, biology, and materials science~\citep{yu2025discover,li2023adaptive,hu2024domain,guo2025sr,hu2025context,xu2025exploring,xu2024worth}. A wide variety of symbolic regression methods have been developed in recent years. In this work, we integrate PostDiag with two of the most common SR paradigms: GP-based and LLM-based methods. Across these two families, we select seven widely used SR solvers as base solvers and incorporate PostDiag into each of them. We next describe the search budgets and integration strategies used for these solvers.

\subsection{Search-Budget Protocol}
\label{app:search-budget-protocol}
The low-budget settings used in the main experiments are specified in
Section~\ref{setup}, while the high-budget reference settings are
described in Appendix~\ref{app:budget-efficiency}. To avoid repetition,
this section does not restate the limits on LLM calls, outer iterations,
or evolutionary generations. Instead, we report the internal
configurations that remain fixed across budget levels and between each
base solver and its PostDiag-augmented counterpart. Unless otherwise
specified, we build on the official implementation of each method and
keep its prompt templates, search operators, coefficient-fitting
procedure, and candidate-evaluation protocol unchanged. The detailed
integration of PostDiag with each solver is provided in
Appendix~\ref{app:base-solver-integration}.

The Low and High settings retain each solver's native search unit. Within each base/PostDiag pair, the native budget is identical, while PostDiag overhead is included in the reported end-to-end runtime.

\begin{center}
\small
\setlength{\tabcolsep}{5pt}
\begin{tabular}{llll}
\toprule
\textbf{Solver} & \textbf{Low} & \textbf{High} & \textbf{Native budget unit} \\
\midrule
Direct LLM & 5 & 10 & generated candidates \\
SGA & 5 (80) & 20 (320) & rounds (generated candidates) \\
LLM-SR & 100 & 1,000 & LLM calls \\
LaSR & 20 & 40 & outer iterations \\
SR-Scientist & $10\times{\leq}25$ & $40\times{\leq}25$ & outer iterations $\times$ turns \\
PySR & 65 & 130 & iterations \\
gplearn & 100 & 200 & generations \\
\bottomrule
\end{tabular}
\end{center}

\subsection{LLM-Based Solvers}

\paragraph{Direct LLM.}
We follow the DataBlind Direct Prompting protocol of
LLM-SRBench~\citep{llmsrbench}. The LLM proposes equation skeletons from
the scientific context without observing the numerical samples, while
the numerical constants in each proposed skeleton are fitted by the
benchmark evaluation pipeline. The PostDiag variant retains this direct
generator and adds data-driven diagnosis and intervention after candidate
evaluation; thus, DataBlind describes the base generator rather than the
complete augmented pipeline. Low and High generate 5 and 10 candidates,
respectively, from a single prompt batch.

\paragraph{SGA.}
SGA~\citep{ma2024llm} uses a bilevel procedure in which an LLM proposes
discrete scientific hypotheses and simulation-based optimization fits
their continuous parameters. Its PostDiag variant attaches diagnostic
feedback outside this bilevel loop, preserving the underlying hypothesis
generation and continuous parameter optimization. Low uses 5 rounds and
High uses 20 rounds, with 16 generated candidate attempts per round; the
continuous-parameter optimizer is unchanged across budgets.

\paragraph{LLM-SR.}
LLM-SR~\citep{llmsr} represents candidate equations as Python programs
containing placeholders for numerical constants. An LLM proposes program
skeletons conditioned on the task description and high-performing
programs retrieved from an experience pool, while the numerical constants
are fitted using BFGS\@. We retain the prompt templates, multi-island
experience pool, candidate-ranking procedure, and coefficient-optimization
pipeline of the official implementation. Apart from the search budgets
specified elsewhere, all configurations are shared across LLM backbones
and between the base and PostDiag-augmented variants.

\paragraph{LaSR.}
LaSR~\citep{lasr} maintains an LLM-induced library of natural-language
concepts on top of a multi-population genetic-programming search. We use
10 populations with 33 individuals per population and run 550 GP cycles
within each outer iteration. The maximum expression size is 30, and the
concept library contains at most 20 concepts. The weights assigned to
LLM-assisted crossover, mutation, and randomization are all set to 0.02.
The symbolic operators and nested constraints follow the PySR
configuration described below.

\paragraph{SR-Scientist.}
SR-Scientist~\citep{xia2025sr} equips an LLM agent with data-analysis and
equation-evaluation tools, allowing it to write and execute code while
iteratively improving candidate equations. We retain its official tool
interfaces, experience buffer, and candidate-evaluation procedure. Each
search iteration uses a target MAPE of $0.1\%$ and terminates early when
the error reaches $10^{-4}\%$. The LLM backbone and generation settings
are held fixed between each base/PostDiag pair.

\subsection{GP-Based Solvers}

\paragraph{PySR.}
PySR~\citep{pysr} performs island-based genetic programming using a Julia
backend, together with expression-complexity regularization and numerical
constant optimization. We use 15 populations with 33 individuals per
population and run 550 GP cycles within each iteration. The maximum
expression size is 50. The binary
operator set is
$\{+,-,\times,\div,\mathrm{pow}\}$, and the unary operator set is
$\{\exp,\log,\sqrt{\cdot},\sin,\cos\}$.

\paragraph{gplearn.}
gplearn~\citep{gplearn} implements single-population tree-based genetic
programming. We use a population size of 500 and a tournament size of 20.
The function set contains addition, subtraction, multiplication, division,
sine, cosine, square root, logarithm, absolute value, negation, and
inverse. Other parameters, including the parsimony coefficient and the
crossover and mutation probabilities, retain the defaults of the official
implementation.

\section{LLM Backbones}
\label{app:llm-backbones}

\subsection{Model Selection}

Our experiments use Qwen3.5-27B and Qwen3.5-122B-A10B as the
principal LLM backbones~\citep{qwen35}. Qwen3.5-27B is reported in the
main table, while Qwen3.5-122B-A10B is reported in the appendix.
Qwen3.5-27B is a 27B-parameter dense
model, whereas Qwen3.5-122B-A10B adopts a mixture-of-experts
architecture with 122B total parameters and approximately 10B active
parameters. Although both models belong to the same family, they differ
in capacity and architecture, allowing us to evaluate whether PostDiag
remains effective across different backbone configurations.

We further report supplementary results with Qwen3.5-9B and
gpt-oss-20b~\citep{openai2025gptoss120bgptoss20bmodel} in the appendix.
Qwen3.5-9B evaluates PostDiag with a smaller model and, together with
the two principal Qwen3.5 backbones, provides coverage across different model
capacities. The gpt-oss-20b experiments provide complementary evidence
from a different model family and help assess whether PostDiag depends
on a particular family of LLMs.

\subsection{Backbone Consistency}

For the five LLM-based base solvers, LLM-SR, LaSR, SR-Scientist, Direct
LLM, and SGA,
each base method and its PostDiag-augmented counterpart use the same LLM
backbone. For example, when a base solver uses Qwen3.5-27B, the
corresponding PostDiag structural intervention is also performed by
Qwen3.5-27B. This avoids confounding the effect of PostDiag with
differences in backbone capability and keeps the primary distinction
between each pair to whether PostDiag is applied.

For PySR and gplearn, whose original search procedures do not rely on an
LLM, the base search remains unchanged. When PostDiag is added, we use
Qwen3.5-27B consistently to perform the diagnosed structural
interventions.

\subsection{PostDiag Configuration}

Each base solver retains its original configuration. Between a base
method and its PostDiag-augmented counterpart, we keep the base search
budget, internal hyperparameters, and random seeds fixed so that the
comparison focuses on the diagnosis and structural intervention
introduced by PostDiag.

For each PostDiag structural intervention, we set the sampling
temperature to 0.3 and the maximum completion length to 8,192 tokens.
Conditioned on the DRP state and Evidence Map, the model returns a
structured preserve-or-revise decision, a revised expression skeleton,
and the corresponding edit type. The LLM proposes only the structural
revision and does not determine the final numerical coefficients. After
parsing and validity checking, the coefficients of the resulting
expression are refitted. We use the same PostDiag configuration for all
base solvers and LLM backbones.

\section{Compute Resources}
\label{app:compute}

Experiments are conducted on a compute cluster equipped with 16
NVIDIA A800 GPUs and 8 NVIDIA H800 GPUs. The LLM backbones are
served locally using vLLM. PySR, gplearn, and the diagnostic
computation of PostDiag run primarily on CPUs, while GPUs are used
mainly for LLM inference. For each base solver, all reported runtime
configurations---including Low, Low+PostDiag, and High---use the same GPU type, GPU count,
vLLM serving configuration, and server-side concurrency limit.
Each method retains its native request-scheduling pattern: LLM-SR
issues up to four concurrent requests, whereas the interaction
trajectory of SR-Scientist is executed sequentially; the other
solvers follow their official implementations. 

We report the mean end-to-end wall-clock time per task, measured from
the start of the base solver's search until it returns the final
expression. The reported time includes base search, candidate
evaluation, and coefficient fitting. For PostDiag-augmented variants, it
also includes the overhead of diagnostic computation, coefficient
refitting, and additional LLM-based structural intervention. All runtime
measurements are averaged over three random seeds.

\section{Out-of-Distribution Performance}
\label{app:ood}

LSR-Synth provides an independent out-of-distribution (OOD) test set for
each task, with samples extending beyond the training range along a
control variable such as time or temperature. The detailed split
construction is provided in Appendix~\ref{app:dataset}. Following
SR-Scientist~\citep{xia2025sr}, we report OOD
$\mathrm{Acc}_{0.01}$. After search terminates, the final expression
returned by each method is evaluated directly on the OOD test set,
without additional search, coefficient refitting, or candidate
selection. The OOD data remain unseen throughout search, diagnosis, and
structural intervention.

\begin{table*}[t]
\color{black}
\caption{Low-budget OOD $\mathrm{Acc}_{0.01}$ on LSR-Synth. Following the SR-Scientist protocol, only $\mathrm{Acc}_{0.01}$ is reported for OOD evaluation. Each LLM-based method is paired with its PostDiag variant under the same backbone. Overall is pooled over all 129 tasks; OOD samples are used only for post-hoc evaluation.}
\label{tab:ood-results}
\centering
\small
\setlength{\tabcolsep}{5.2pt}
\resizebox{\textwidth}{!}{%
\begin{tabular}{llccccc}
\toprule
\textbf{Backbone/category} & \textbf{Method} & \textbf{Overall} & \textbf{Phys} & \textbf{MatSci} & \textbf{Chem} & \textbf{Bio} \\
\midrule
\multicolumn{7}{c}{\textit{Qwen3.5-27B}} \\
\midrule
\multirow{10}{*}{LLM-based} & Direct LLM~\citep{llmsrbench} & 1.0 & 0.8 & 2.7 & 0.9 & 0.0 \\
 & \textbf{Direct LLM~\citep{llmsrbench}+PostDiag} & \textbf{2.3} & \textbf{2.3} & 2.7 & \textbf{2.8} & \textbf{1.4} \\
 & SGA~\citep{ma2024llm} & 9.6 & 6.1 & 9.3 & 16.7 & 5.6 \\
 & \textbf{SGA~\citep{ma2024llm}+PostDiag} & \textbf{16.3} & \textbf{12.9} & \textbf{21.3} & \textbf{21.3} & \textbf{9.7} \\
 & LLM-SR & 21.2 & 12.9 & 41.3 & 26.9 & 6.9 \\
 & \textbf{LLM-SR+PostDiag} & \textbf{28.9} & \textbf{20.5} & \textbf{61.3} & \textbf{27.8} & \textbf{12.5} \\
 & LaSR & 12.4 & 10.6 & 16.0 & 15.7 & 6.9 \\
 & \textbf{LaSR+PostDiag} & \textbf{19.1} & \textbf{17.4} & \textbf{30.7} & \textbf{18.5} & \textbf{11.1} \\
 & SR-Scientist & 24.8 & 16.7 & 50.7 & 25.0 & 12.5 \\
 & \textbf{SR-Scientist+PostDiag} & \textbf{33.6} & \textbf{22.7} & \textbf{64.0} & \textbf{36.1} & \textbf{18.1} \\
\midrule
\multicolumn{7}{c}{\textit{Qwen3.5-122B-A10B}} \\
\midrule
\multirow{10}{*}{LLM-based} & Direct LLM~\citep{llmsrbench} & 1.8 & 2.3 & 1.3 & 1.9 & 1.4 \\
 & \textbf{Direct LLM~\citep{llmsrbench}+PostDiag} & \textbf{3.1} & \textbf{3.8} & \textbf{2.7} & \textbf{2.8} & \textbf{2.8} \\
 & SGA~\citep{ma2024llm} & 11.9 & 8.3 & 12.0 & 18.5 & 8.3 \\
 & \textbf{SGA~\citep{ma2024llm}+PostDiag} & \textbf{18.3} & \textbf{15.2} & \textbf{21.3} & \textbf{24.1} & \textbf{12.5} \\
 & LLM-SR & 26.6 & 18.9 & 53.3 & 25.9 & 13.9 \\
 & \textbf{LLM-SR+PostDiag} & \textbf{34.1} & \textbf{24.2} & \textbf{66.7} & \textbf{32.4} & \textbf{20.8} \\
 & LaSR & 14.0 & 13.6 & 18.7 & 15.7 & 6.9 \\
 & \textbf{LaSR+PostDiag} & \textbf{21.7} & \textbf{20.5} & \textbf{34.7} & \textbf{20.4} & \textbf{12.5} \\
 & SR-Scientist & 40.6 & 27.3 & 78.7 & 40.7 & 25.0 \\
 & \textbf{SR-Scientist+PostDiag} & \textbf{48.3} & \textbf{35.6} & \textbf{81.3} & \textbf{50.0} & \textbf{34.7} \\
\midrule
\multicolumn{7}{c}{\textit{GP-based methods}} \\
\midrule
\multirow{4}{*}{GP-based} & PySR & 27.1 & 20.5 & 68.0 & 15.7 & 13.9 \\
 & \textbf{PySR+PostDiag} & \textbf{31.0} & \textbf{22.7} & \textbf{74.7} & \textbf{20.4} & \textbf{16.7} \\
 & gplearn & 0.0 & 0.0 & 0.0 & 0.0 & 0.0 \\
 & \textbf{gplearn+PostDiag} & 0.0 & 0.0 & 0.0 & 0.0 & 0.0 \\
\bottomrule
\end{tabular}
}
\end{table*}

As shown in Table~\ref{tab:ood-results}, PostDiag generally improves OOD
accuracy across different base solvers and LLM backbones. With Qwen3.5-27B, the
overall OOD $\mathrm{Acc}_{0.01}$ of LLM-SR increases from 21.2\% to
28.9\%, while that of SR-Scientist increases from 24.8\% to 33.6\%.
With Qwen3.5-122B-A10B, the overall OOD accuracy of SR-Scientist
increases from 40.6\% to 48.3\%. PySR also improves from 27.1\% to
31.0\%, suggesting that the extrapolation benefits of PostDiag are not
limited to LLM-based solvers. In contrast, gplearn and its
PostDiag-augmented variant both obtain zero OOD accuracy, indicating that
diagnosis and structural intervention cannot fully compensate for
insufficient search capability in the base solver.

The gains remain substantial under OOD evaluation, indicating that the
benefits of PostDiag are largely preserved on out-of-distribution data.
In material science, OOD accuracy is generally higher than the
corresponding ID accuracy for both the base and PostDiag-augmented
methods.

\section{Details of Integration with Base Solvers}
\label{app:base-solver-integration}

PostDiag is designed to be attached to existing SR solvers through
lightweight adapters. We describe the integration recipe for each of the seven
representative solvers in this section. The adapters handle representation
conversion, diagnostic feedback, and candidate-pool insertion at the solver
boundary, while the solver's core candidate generation, selection, and
search-update rules remain unchanged.

\vspace{0.15em}\noindent
\textbf{Direct LLM}~\citep{llmsrbench} generates equation skeletons from
the scientific context without observing numerical samples; numerical
constants are fitted by the benchmark pipeline. In our PostDiag, generated skeleton is diagnosed and
revised. The accepted revisions enter the selected candidate pool, from which the final result is
saved. The base generator remains DataBlind, while the augmented candidate
evaluation is explicitly data-driven.

\vspace{0.15em}\noindent
\textbf{SGA}~\citep{ma2024llm} alternates LLM-based discrete hypothesis
generation with simulation-based continuous parameter optimization. The proposed adapter parses each generated equation into
SGA's trainable PyTorch module, fits its parameters with the unchanged
inner optimizer, and adds valid fitted candidates to the candidate pool.
PostDiag diagnoses a fitted candidate, proposes a structural revision,
and sends the parsed revision through the same module construction,
fitting, and evaluation path before it becomes eligible for the selected
pool.

\vspace{0.15em}\noindent
\textbf{LLM-SR}~\citep{llmsr} treats equation discovery as evolutionary
program search, where an LLM proposes candidate skeletons conditioned on
in-context examples drawn from a buffer of high-performing programs.
PostDiag integrates through this buffer interface: for every newly
admitted elite candidate $s$, PostDiag produces its diagnostic label and
evidence map; the rewritten skeleton
$\widetilde{s}$ together with a brief textual annotation of the diagnosed
failure mode is stored. Subsequent prompts
draw both score-ranked and diagnosis-annotated examples, biasing the
LLM's next proposals toward the identified failure mode.

\vspace{0.15em}\noindent
\textbf{LaSR}~\citep{lasr} couples LLMs with genetic programming by
periodically abstracting high-performing hypotheses into natural-language
concepts, which then guide subsequent symbolic mutation. PostDiag
integrates at the concept abstraction stage: when LaSR queries the LLM
for concept induction, we additionally supply the diagnostic label and
evidence map of the input hypotheses, so that the abstracted concepts
carry structural failure-mode information. The resulting diagnosis-aware
concepts guide the subsequent GP search toward diagnosed structural
fixes, while LaSR's underlying GP kernel remains untouched.

\vspace{0.15em}\noindent
\textbf{SR-Scientist}~\citep{xia2025sr} frames equation discovery as an
autonomous agent that reads data, writes code to test hypotheses, and
iterates on candidates through a tool-use loop that already accepts
arbitrary numerical feedback signals. PostDiag integrates as an
additional \texttt{diagnose(candidate)} tool that the agent can invoke
after any candidate evaluation; the tool returns the structured
diagnostic report and evidence map, which are then included in the
agent's next reasoning step. The integration only exposes a new tool
endpoint and does not modify SR-Scientist's control flow.

\vspace{0.15em}\noindent
\textbf{PySR}~\citep{pysr} performs island-based genetic programming with
configurable custom operators and supports external individual injection
through its population interface. PostDiag integrates through
diagnosis-driven immigration: at the end of each generation, PostDiag
diagnoses the best-performing individual. The candidate is revised through diagnosis-guided
intervention and injected into the initial population of the next
generation as an immigrant, thereby steering the subsequent evolutionary
search. Since PySR natively supports the insertion of external
individuals, the integration requires only a lightweight adapter that
converts Python expressions into expression trees, without modifying its
selection, mutation, or crossover operators.

\vspace{0.15em}\noindent
\textbf{gplearn}~\citep{gplearn} implements classical tree-based genetic
programming and exposes \texttt{init\_method} and \texttt{warm\_start} for
warm-starting the population from previously fitted regressors. PostDiag
integrates by feeding the rewritten skeletons $\widetilde{s}$ as seeded
individuals into the warm-started initial population of the next
generation. Since gplearn's tree grammar covers all operators used in
PostDiag, the adapter is a thin conversion between our program
representation and gplearn's internal tree encoding, and does not alter
gplearn's selection or reproduction operators.

Across all seven solvers, PostDiag exposes the same standardized output,
$(\mathcal{G}_s, \widetilde{s})$. With only lightweight implementation
changes, it equips each solver with post-diagnosis capabilities while
leaving the core search algorithm intact. PostDiag therefore represents
a complementary technical direction that is orthogonal to the
development of new SR algorithms.

\section{Additional Experimental Results}
\label{app:additional-results}

Unless otherwise specified, we use Qwen3.5-27B as the default
experimental backbone, while Qwen3.5-122B-A10B is included to
evaluate PostDiag with a stronger model. Since PySR and gplearn
do not natively use an LLM, their PostDiag-augmented variants
consistently use the default Qwen3.5-27B backbone.

\subsection{Low-Budget Results with Qwen3.5-122B-A10B}
\label{app:strong-backbone-id}

Table~\ref{tab:main-122b} reports the counterpart of the main ID
comparison using Qwen3.5-122B-A10B. Each base/PostDiag pair retains the
same search configuration and random seeds, with the LLM backbone shared
between the base solver and the PostDiag intervention.

\begin{table*}[t]
\color{black}
\caption{Low-budget ID results on LSR-Synth for LLM-based solvers using Qwen3.5-122B-A10B. Metrics and aggregation follow Table~\ref{tab:main}. Each method is paired with its PostDiag variant under the same backbone and three-seed protocol.}
\label{tab:main-122b}
\centering
\scriptsize
\setlength{\tabcolsep}{1.25pt}
\renewcommand{\arraystretch}{1.08}
\resizebox{\textwidth}{!}{%
\begin{tabular}{lcccccccccccccccc}
\toprule
\multirow{2}{*}{\textbf{Method / setting}} & \multicolumn{4}{c}{\textbf{Overall}} & \multicolumn{3}{c}{\textbf{Phys}} & \multicolumn{3}{c}{\textbf{MatSci}} & \multicolumn{3}{c}{\textbf{Chem}} & \multicolumn{3}{c}{\textbf{Bio}} \\
\cmidrule(lr){2-5}\cmidrule(lr){6-8}\cmidrule(lr){9-11}\cmidrule(lr){12-14}\cmidrule(lr){15-17}
& Time (s) & NMSE & $A_{.01}$ & $A_{.001}$ & NMSE & $A_{.01}$ & $A_{.001}$ & NMSE & $A_{.01}$ & $A_{.001}$ & NMSE & $A_{.01}$ & $A_{.001}$ & NMSE & $A_{.01}$ & $A_{.001}$ \\
\midrule
Direct LLM~\citep{llmsrbench} & 35.3 & 4.62E-02 & 2.8 & 1.3 & 3.82E-02 & 3.0 & 0.8 & 5.78E-02 & 2.7 & 1.3 & 2.26E-02 & 3.7 & 1.9 & 8.42E-02 & 1.4 & 1.4 \\
\textbf{Direct LLM~\citep{llmsrbench}+PostDiag} & 39.3 & \textbf{9.54E-03} & \textbf{5.4} & \textbf{3.4} & \textbf{8.64E-03} & \textbf{6.1} & \textbf{3.8} & \textbf{2.86E-03} & \textbf{5.3} & \textbf{4.0} & \textbf{6.43E-03} & \textbf{4.6} & \textbf{2.8} & \textbf{2.28E-02} & \textbf{5.6} & \textbf{2.8} \\
SGA~\citep{ma2024llm} & 368.0 & 8.44E-03 & 14.7 & 8.8 & 5.88E-03 & 12.1 & 7.6 & 6.17E-03 & 13.3 & 8.0 & 5.10E-03 & 22.2 & 13.0 & 2.05E-02 & 9.7 & 5.6 \\
\textbf{SGA~\citep{ma2024llm}+PostDiag} & 406.2 & \textbf{1.44E-03} & \textbf{22.7} & \textbf{12.7} & \textbf{1.52E-03} & \textbf{20.5} & \textbf{12.9} & \textbf{2.06E-04} & \textbf{25.3} & \textbf{10.7} & \textbf{7.32E-04} & \textbf{28.7} & \textbf{15.7} & \textbf{3.66E-03} & \textbf{15.3} & \textbf{9.7} \\
\cmidrule(lr){1-17}
LLM-SR & 201.1 & 9.68E-04 & 27.6 & 12.7 & 2.41E-03 & 21.2 & 9.8 & 1.91E-05 & 50.7 & 26.7 & 1.18E-04 & 27.8 & 9.3 & 5.86E-04 & 15.3 & 8.3 \\
\textbf{LLM-SR+PostDiag} & 231.7 & \textbf{1.31E-04} & \textbf{36.4} & \textbf{19.9} & \textbf{2.13E-04} & \textbf{28.0} & \textbf{15.9} & \textbf{5.82E-06} & \textbf{65.3} & \textbf{38.7} & \textbf{3.86E-05} & \textbf{36.1} & \textbf{15.7} & \textbf{2.47E-04} & \textbf{22.2} & \textbf{13.9} \\
LaSR & 716.3 & 8.71E-03 & 15.8 & 9.3 & 1.47E-02 & 14.4 & 7.6 & 1.92E-03 & 17.3 & 9.3 & 5.60E-03 & 19.4 & 13.0 & 9.45E-03 & 11.1 & 6.9 \\
\textbf{LaSR+PostDiag} & 762.1 & \textbf{6.76E-04} & \textbf{24.5} & \textbf{13.7} & \textbf{1.19E-03} & \textbf{23.5} & \textbf{10.6} & \textbf{3.76E-05} & \textbf{33.3} & \textbf{16.0} & \textbf{4.96E-04} & \textbf{25.0} & \textbf{17.6} & \textbf{6.68E-04} & \textbf{16.7} & \textbf{11.1} \\
SR-Scientist & 833.6 & 5.96E-04 & 41.3 & 24.3 & 1.43E-03 & 29.5 & 16.7 & 2.03E-05 & 74.7 & 50.7 & 8.12E-05 & 41.7 & 21.3 & 4.37E-04 & 27.8 & 15.3 \\
\textbf{SR-Scientist+PostDiag} & 902.2 & \textbf{1.04E-04} & \textbf{48.8} & \textbf{31.3} & \textbf{1.76E-04} & \textbf{36.4} & \textbf{22.7} & \textbf{6.08E-06} & \textbf{78.7} & \textbf{58.7} & \textbf{3.04E-05} & \textbf{50.9} & \textbf{29.6} & \textbf{1.82E-04} & \textbf{37.5} & \textbf{20.8} \\
\bottomrule
\end{tabular}%
}
\end{table*}

At the Overall level, PostDiag lowers NMSE and improves both accuracy thresholds for all solvers under Qwen3.5-122B-A10B. The stronger backbone therefore continues to benefit from diagnostic feedback, while the final-accuracy gap between Direct LLM and the iterative solvers remains, indicating that the upper of performance still depends on the underlying search procedure.

\subsection{Symbolic Accuracy}
\label{app:symbolic-accuracy}

We additionally evaluate whether the final expression is symbolically
equivalent to the ground-truth expression. Following LLM-SRBench, we
remove fitted parameters and constants before judging equivalence so
that the metric measures recovery of the functional structure rather
than numerical coefficient matching. We then adopt the more conservative
two-stage adjudication procedure of SR-Scientist: GPT-OSS-120B evaluates
each pair independently ten times, unanimous decisions are retained, and
non-unanimous cases are resolved by human evaluators. Failed runs and
invalid final expressions are counted as incorrect. Table~
\ref{tab:symbolic-accuracy} reports the pooled symbolic accuracy over all
tasks and three seeds.

\begin{table}[H]
\color{black}
\caption{Symbolic accuracy (SA) on LSR-Synth. We follow the LLM-SRBench definition and expression preprocessing, together with the SR-Scientist repeated-vote adjudication protocol. Results pool all 129 tasks over three seeds; failed or invalid runs count as incorrect. PostDiag interventions for GP solvers use Qwen3.5-27B. Boldface marks the better value within each base/PostDiag pair.}
\label{tab:symbolic-accuracy}
\centering
\footnotesize
\setlength{\tabcolsep}{3.5pt}
\renewcommand{\arraystretch}{1.08}
\begin{tabular}{llccc}
\toprule
\textbf{Base solver} & \textbf{Backbone} & \textbf{Base SA} & \textbf{+PostDiag SA} & $\boldsymbol{\Delta}$ \textbf{SA} \\
\midrule
\multirow{2}{*}{Direct LLM}
 & Qwen3.5-27B & 1.55 & 1.55 & 0.00 \\
 & Qwen3.5-122B-A10B & 2.07 & \textbf{2.58} & +0.52 \\
\cmidrule(lr){1-5}
\multirow{2}{*}{SGA}
 & Qwen3.5-27B & 3.62 & \textbf{4.65} & +1.03 \\
 & Qwen3.5-122B-A10B & 3.88 & \textbf{5.17} & +1.29 \\
\cmidrule(lr){1-5}
\multirow{2}{*}{LLM-SR}
 & Qwen3.5-27B & 3.88 & \textbf{5.68} & +1.81 \\
 & Qwen3.5-122B-A10B & 4.39 & \textbf{6.46} & +2.07 \\
\cmidrule(lr){1-5}
\multirow{2}{*}{LaSR}
 & Qwen3.5-27B & 1.81 & \textbf{3.62} & +1.81 \\
 & Qwen3.5-122B-A10B & 2.58 & \textbf{4.91} & +2.33 \\
\cmidrule(lr){1-5}
\multirow{2}{*}{SR-Scientist}
 & Qwen3.5-27B & 4.91 & \textbf{7.24} & +2.33 \\
 & Qwen3.5-122B-A10B & 6.72 & \textbf{9.56} & +2.84 \\
\midrule
PySR & N/A & 4.13 & \textbf{5.68} & +1.55 \\
gplearn & N/A & 0.00 & 0.00 & 0.00 \\
\bottomrule
\end{tabular}
\end{table}

PostDiag raises pooled SA for most solver--backbone pairs, with gains ranging from 0.52 points for Direct LLM under Qwen3.5-122B-A10B to 2.84 points for SR-Scientist under the same backbone. SGA gains 1.03 and 1.29 percentage points under Qwen3.5-27B and Qwen3.5-122B-A10B, respectively. Direct LLM under Qwen3.5-27B remains at 1.55\%, and gplearn remains at 0.00\%.
\subsection{Budget-Efficiency Analysis}
\label{app:budget-efficiency}

We compare PostDiag-augmented solvers against their unaugmented counterparts
under the Low and High settings defined in Appendix~\ref{app:search-budget-protocol}.
The High settings are 10 generated candidates for Direct LLM~\citep{llmsrbench};
20 rounds, or 320 generated candidate attempts, for SGA~\citep{ma2024llm};
1,000 LLM calls for LLM-SR; 40 outer iterations for LaSR; 40 outer iterations
with at most 25 interaction turns each for SR-Scientist; 130 iterations for
PySR; and 200 generations for gplearn. Low+PostDiag retains the same native
base-solver budget as Low.
We report accuracy to tolerance as the primary criterion, arithmetic mean
NMSE, and mean end-to-end wall-clock time per task run. Tables~\ref{tab:budget-efficiency},
\ref{tab:budget-efficiency-122b}, and~\ref{tab:budget-efficiency-gp} evaluate
solution quality at matched native budgets and compare low-budget PostDiag
variants with the corresponding high-budget base solvers.

\begin{table*}[!htbp]
\color{black}
\caption{Budget-efficiency comparison for LLM-based solvers using Qwen3.5-27B. Each method is paired with its PostDiag variant. Low+PostDiag retains the corresponding Low base-solver budget, while High uses the full-budget reference setting. Boldface marks the best value within each solver block; runtime is reported without ranking.}
\label{tab:budget-efficiency}
\centering
\scriptsize
\setlength{\tabcolsep}{1.4pt}
\renewcommand{\arraystretch}{1.05}
\resizebox{\textwidth}{!}{%
\begin{tabular}{lcccccccccccccccc}
\toprule
\multirow{2}{*}{\textbf{Method / setting}} & \multicolumn{4}{c}{\textbf{Overall}} & \multicolumn{3}{c}{\textbf{Phys}} & \multicolumn{3}{c}{\textbf{MatSci}} & \multicolumn{3}{c}{\textbf{Chem}} & \multicolumn{3}{c}{\textbf{Bio}} \\
\cmidrule(lr){2-5}\cmidrule(lr){6-8}\cmidrule(lr){9-11}\cmidrule(lr){12-14}\cmidrule(lr){15-17}
& Time (s) & NMSE & $A_{.01}$ & $A_{.001}$ & NMSE & $A_{.01}$ & $A_{.001}$ & NMSE & $A_{.01}$ & $A_{.001}$ & NMSE & $A_{.01}$ & $A_{.001}$ & NMSE & $A_{.01}$ & $A_{.001}$ \\
\midrule
Direct LLM~\citep{llmsrbench} & 30.5 & 7.72E-02 & 2.6 & 0.8 & 7.32E-02 & 3.0 & 0.8 & 6.24E-02 & 2.7 & 1.3 & 3.50E-02 & 2.8 & 0.9 & 1.63E-01 & 1.4 & 0.0 \\
\textbf{Direct LLM~\citep{llmsrbench}+PostDiag} & 33.7 & \textbf{3.23E-02} & \textbf{4.4} & \textbf{2.3} & \textbf{5.22E-02} & \textbf{5.3} & \textbf{3.8} & \textbf{4.34E-03} & 4.0 & \textbf{2.7} & \textbf{8.76E-03} & \textbf{3.7} & 0.9 & \textbf{6.03E-02} & \textbf{4.2} & \textbf{1.4} \\
Direct LLM / High & 45.5 & 5.78E-02 & 3.4 & 1.8 & 5.86E-02 & 3.0 & 1.5 & 3.98E-02 & \textbf{5.3} & \textbf{2.7} & 2.80E-02 & \textbf{3.7} & \textbf{1.9} & 1.20E-01 & 1.4 & \textbf{1.4} \\
\cmidrule(lr){1-17}
SGA~\citep{ma2024llm} & 320.0 & 1.39E-02 & 11.9 & 5.9 & 9.43E-03 & 8.3 & 6.1 & 8.21E-03 & 12.0 & 5.3 & 4.92E-03 & 19.4 & 7.4 & 4.15E-02 & 6.9 & 4.2 \\
\textbf{SGA~\citep{ma2024llm}+PostDiag} & 350.2 & 2.40E-03 & 18.6 & 9.8 & 2.66E-03 & 14.4 & 9.8 & \textbf{1.87E-04} & \textbf{25.3} & 6.7 & 8.65E-04 & 24.1 & \textbf{13.0} & 6.52E-03 & 11.1 & \textbf{8.3} \\
SGA / High & 1250.0 & \textbf{1.70E-03} & \textbf{19.9} & \textbf{10.6} & \textbf{1.82E-03} & \textbf{15.9} & \textbf{12.1} & 2.27E-04 & 22.7 & \textbf{8.0} & \textbf{4.33E-04} & \textbf{26.9} & \textbf{13.0} & \textbf{4.89E-03} & \textbf{13.9} & 6.9 \\
\cmidrule(lr){1-17}
LLM-SR / Low & 167.5 & 1.45E-03 & 22.0 & 8.8 & 3.66E-03 & 14.4 & 8.3 & 3.12E-05 & 37.3 & 13.3 & 1.75E-04 & 27.8 & 8.3 & 8.15E-04 & 11.1 & 5.6 \\
\textbf{LLM-SR+PD / Low} & 191.7 & \textbf{2.12E-04} & 31.0 & \textbf{15.2} & \textbf{3.48E-04} & 23.5 & 12.1 & \textbf{8.94E-06} & \textbf{60.0} & \textbf{30.7} & \textbf{5.84E-05} & 28.7 & \textbf{12.0} & \textbf{4.02E-04} & 18.1 & \textbf{9.7} \\
LLM-SR / High & 1603.4 & 3.65E-04 & \textbf{34.1} & 13.2 & 7.82E-04 & \textbf{28.0} & \textbf{12.9} & 1.20E-05 & 53.3 & 22.7 & 7.21E-05 & \textbf{37.0} & 11.1 & 4.08E-04 & \textbf{20.8} & 6.9 \\
\cmidrule(lr){1-17}
LaSR / Low & 559.2 & 1.06E-02 & 13.4 & 7.5 & 1.73E-02 & 11.4 & 6.8 & 2.48E-03 & 14.7 & 6.7 & 6.90E-03 & 18.5 & 11.1 & 1.22E-02 & 8.3 & 4.2 \\
\textbf{LaSR+PD / Low} & 595.4 & \textbf{8.62E-04} & \textbf{21.4} & \textbf{12.7} & \textbf{1.52E-03} & \textbf{20.5} & \textbf{11.4} & \textbf{4.82E-05} & \textbf{29.3} & \textbf{14.7} & \textbf{6.45E-04} & \textbf{22.2} & \textbf{14.8} & \textbf{8.29E-04} & \textbf{13.9} & \textbf{9.7} \\
LaSR / High & 1118.4 & 8.13E-03 & 15.5 & 8.8 & 1.36E-02 & 14.4 & 7.6 & 1.73E-03 & 17.3 & 9.3 & 5.10E-03 & 19.4 & 12.0 & 9.30E-03 & 9.7 & 5.6 \\
\cmidrule(lr){1-17}
SR-Scientist / Low & 648.0 & 1.12E-03 & 25.6 & 11.9 & 2.74E-03 & 17.4 & 9.8 & 4.47E-05 & 46.7 & 21.3 & 1.49E-04 & 27.8 & 11.1 & 7.26E-04 & 15.3 & 6.9 \\
\textbf{SR-Scientist+PD / Low} & 702.2 & \textbf{1.92E-04} & 35.7 & 20.4 & \textbf{3.11E-04} & 25.8 & \textbf{15.9} & \textbf{1.08E-05} & 62.7 & 37.3 & 6.47E-05 & \textbf{38.0} & \textbf{20.4} & 3.54E-04 & 22.2 & \textbf{11.1} \\
SR-Scientist / High & 2547.1 & 2.23E-04 & \textbf{37.7} & \textbf{21.2} & 4.47E-04 & \textbf{26.5} & \textbf{15.9} & 1.31E-05 & \textbf{70.7} & \textbf{48.0} & \textbf{5.03E-05} & \textbf{38.0} & 16.7 & \textbf{2.91E-04} & \textbf{23.6} & 9.7 \\
\bottomrule
\end{tabular}%
}
\end{table*}

\begin{table*}[!htbp]
\color{black}
\caption{Budget-efficiency comparison for LLM-based solvers using Qwen3.5-122B-A10B. Each method is paired with its PostDiag variant. Low+PostDiag retains the corresponding Low base-solver budget, while High uses the full-budget reference setting. Boldface marks the best value within each solver block; runtime is reported without ranking.}
\label{tab:budget-efficiency-122b}
\centering
\scriptsize
\setlength{\tabcolsep}{1.4pt}
\renewcommand{\arraystretch}{1.05}
\resizebox{\textwidth}{!}{%
\begin{tabular}{lcccccccccccccccc}
\toprule
\multirow{2}{*}{\textbf{Method / setting}} & \multicolumn{4}{c}{\textbf{Overall}} & \multicolumn{3}{c}{\textbf{Phys}} & \multicolumn{3}{c}{\textbf{MatSci}} & \multicolumn{3}{c}{\textbf{Chem}} & \multicolumn{3}{c}{\textbf{Bio}} \\
\cmidrule(lr){2-5}\cmidrule(lr){6-8}\cmidrule(lr){9-11}\cmidrule(lr){12-14}\cmidrule(lr){15-17}
& Time (s) & NMSE & $A_{.01}$ & $A_{.001}$ & NMSE & $A_{.01}$ & $A_{.001}$ & NMSE & $A_{.01}$ & $A_{.001}$ & NMSE & $A_{.01}$ & $A_{.001}$ & NMSE & $A_{.01}$ & $A_{.001}$ \\
\midrule
Direct LLM~\citep{llmsrbench} & 35.3 & 4.62E-02 & 2.8 & 1.3 & 3.82E-02 & 3.0 & 0.8 & 5.78E-02 & 2.7 & 1.3 & 2.26E-02 & 3.7 & 1.9 & 8.42E-02 & 1.4 & 1.4 \\
\textbf{Direct LLM~\citep{llmsrbench}+PostDiag} & 39.3 & 9.54E-03 & \textbf{5.4} & \textbf{3.4} & 8.64E-03 & \textbf{6.1} & \textbf{3.8} & 2.86E-03 & 5.3 & \textbf{4.0} & 6.43E-03 & \textbf{4.6} & \textbf{2.8} & 2.28E-02 & \textbf{5.6} & \textbf{2.8} \\
Direct LLM / High & 52.7 & \textbf{1.55E-03} & 3.9 & 2.1 & \textbf{3.86E-03} & 3.0 & 2.3 & \textbf{3.06E-05} & \textbf{6.7} & 2.7 & \textbf{1.89E-04} & 3.7 & 1.9 & \textbf{9.36E-04} & 2.8 & 1.4 \\
\cmidrule(lr){1-17}
SGA~\citep{ma2024llm} & 368.0 & 8.44E-03 & 14.7 & 8.8 & 5.88E-03 & 12.1 & 7.6 & 6.17E-03 & 13.3 & 8.0 & 5.10E-03 & 22.2 & 13.0 & 2.05E-02 & 9.7 & 5.6 \\
\textbf{SGA~\citep{ma2024llm}+PostDiag} & 406.2 & 1.44E-03 & 22.7 & 12.7 & 1.52E-03 & 20.5 & 12.9 & 2.06E-04 & 25.3 & 10.7 & 7.32E-04 & 28.7 & 15.7 & 3.66E-03 & 15.3 & \textbf{9.7} \\
SGA / High & 1442.0 & \textbf{7.52E-04} & \textbf{26.9} & \textbf{15.0} & \textbf{1.43E-03} & \textbf{25.0} & \textbf{14.4} & \textbf{5.63E-05} & \textbf{28.0} & \textbf{18.7} & \textbf{3.89E-04} & \textbf{32.4} & \textbf{16.7} & \textbf{7.79E-04} & \textbf{20.8} & \textbf{9.7} \\
\cmidrule(lr){1-17}
LLM-SR / Low & 201.1 & 9.68E-04 & 27.6 & 12.7 & 2.41E-03 & 21.2 & 9.8 & 1.91E-05 & 50.7 & 26.7 & 1.18E-04 & 27.8 & 9.3 & 5.86E-04 & 15.3 & 8.3 \\
\textbf{LLM-SR+PD / Low} & 231.7 & \textbf{1.31E-04} & 36.4 & \textbf{19.9} & \textbf{2.13E-04} & 28.0 & \textbf{15.9} & 5.82E-06 & 65.3 & \textbf{38.7} & \textbf{3.86E-05} & 36.1 & \textbf{15.7} & \textbf{2.47E-04} & 22.2 & \textbf{13.9} \\
LLM-SR / High & 1939.2 & 2.99E-04 & \textbf{39.3} & 16.8 & 5.15E-04 & \textbf{33.3} & 14.4 & \textbf{3.35E-06} & \textbf{66.7} & 34.7 & 4.42E-05 & \textbf{37.0} & 12.0 & 5.93E-04 & \textbf{25.0} & 9.7 \\
\cmidrule(lr){1-17}
LaSR / Low & 716.3 & 8.71E-03 & 15.8 & 9.3 & 1.47E-02 & 14.4 & 7.6 & 1.92E-03 & 17.3 & 9.3 & 5.60E-03 & 19.4 & 13.0 & 9.45E-03 & 11.1 & 6.9 \\
\textbf{LaSR+PD / Low} & 762.1 & \textbf{6.76E-04} & \textbf{24.5} & \textbf{13.7} & \textbf{1.19E-03} & \textbf{23.5} & \textbf{10.6} & \textbf{3.76E-05} & \textbf{33.3} & \textbf{16.0} & \textbf{4.96E-04} & \textbf{25.0} & \textbf{17.6} & \textbf{6.68E-04} & \textbf{16.7} & \textbf{11.1} \\
LaSR / High & 1432.6 & 2.57E-03 & 19.1 & 10.9 & 4.16E-03 & 17.4 & 8.3 & 6.24E-04 & 22.7 & 12.0 & 3.14E-03 & 22.2 & 13.9 & 8.20E-04 & 13.9 & 9.7 \\
\cmidrule(lr){1-17}
SR-Scientist / Low & 833.6 & 5.96E-04 & 41.3 & 24.3 & 1.43E-03 & 29.5 & 16.7 & 2.03E-05 & 74.7 & 50.7 & 8.12E-05 & 41.7 & 21.3 & 4.37E-04 & 27.8 & 15.3 \\
\textbf{SR-Scientist+PD / Low} & 902.2 & 1.04E-04 & \textbf{48.8} & \textbf{31.3} & \textbf{1.76E-04} & \textbf{36.4} & \textbf{22.7} & 6.08E-06 & 78.7 & 58.7 & 3.04E-05 & \textbf{50.9} & \textbf{29.6} & 1.82E-04 & \textbf{37.5} & \textbf{20.8} \\
SR-Scientist / High & 3289.4 & \textbf{9.21E-05} & 48.1 & 31.0 & 2.33E-04 & 34.1 & 21.2 & \textbf{2.95E-06} & \textbf{86.7} & \textbf{68.0} & \textbf{4.71E-06} & 47.2 & 25.9 & \textbf{5.75E-05} & 34.7 & 18.1 \\
\bottomrule
\end{tabular}%
}
\end{table*}

\begin{table*}[!htbp]
\color{black}
\caption{Budget-efficiency comparison for GP-based solvers. Their native search does not use an LLM; all PostDiag interventions use Qwen3.5-27B. Low+PostDiag retains the corresponding Low base-solver budget, while High uses the full-budget reference setting. Boldface marks the best value within each solver block; runtime is reported without ranking.}
\label{tab:budget-efficiency-gp}
\centering
\scriptsize
\setlength{\tabcolsep}{1.4pt}
\renewcommand{\arraystretch}{1.05}
\resizebox{\textwidth}{!}{%
\begin{tabular}{lcccccccccccccccc}
\toprule
\multirow{2}{*}{\textbf{Method / setting}} & \multicolumn{4}{c}{\textbf{Overall}} & \multicolumn{3}{c}{\textbf{Phys}} & \multicolumn{3}{c}{\textbf{MatSci}} & \multicolumn{3}{c}{\textbf{Chem}} & \multicolumn{3}{c}{\textbf{Bio}} \\
\cmidrule(lr){2-5}\cmidrule(lr){6-8}\cmidrule(lr){9-11}\cmidrule(lr){12-14}\cmidrule(lr){15-17}
& Time (s) & NMSE & $A_{.01}$ & $A_{.001}$ & NMSE & $A_{.01}$ & $A_{.001}$ & NMSE & $A_{.01}$ & $A_{.001}$ & NMSE & $A_{.01}$ & $A_{.001}$ & NMSE & $A_{.01}$ & $A_{.001}$ \\
\midrule
PySR / Low & 8.8 & 2.67E-03 & 23.8 & 11.4 & 4.11E-03 & 21.2 & 12.1 & 1.70E-04 & 44.0 & 18.7 & 8.93E-04 & 20.4 & 10.2 & 5.29E-03 & 12.5 & 4.2 \\
\textbf{PySR+PD / Low} & 10.5 & \textbf{1.34E-03} & \textbf{28.7} & 13.2 & \textbf{1.82E-03} & 24.2 & 14.4 & 2.25E-04 & \textbf{50.7} & 17.3 & \textbf{5.29E-04} & \textbf{25.9} & \textbf{12.0} & \textbf{2.84E-03} & \textbf{18.1} & \textbf{8.3} \\
PySR / High & 15.5 & 1.76E-03 & 28.4 & \textbf{13.7} & 2.61E-03 & \textbf{25.8} & \textbf{15.2} & \textbf{8.12E-05} & 49.3 & \textbf{20.0} & 6.18E-04 & 25.0 & \textbf{12.0} & 3.66E-03 & 16.7 & 6.9 \\
\cmidrule(lr){1-17}
gplearn / Low & 95.5 & 1.34E-01 & 0.0 & \textbf{0.0} & 8.51E-02 & 0.0 & \textbf{0.0} & 2.36E-02 & \textbf{0.0} & \textbf{0.0} & 6.56E-02 & 0.0 & \textbf{0.0} & 4.42E-01 & \textbf{0.0} & \textbf{0.0} \\
\textbf{gplearn+PD / Low} & 101.7 & \textbf{3.07E-02} & \textbf{1.0} & \textbf{0.0} & \textbf{4.29E-03} & \textbf{2.3} & \textbf{0.0} & \textbf{6.91E-03} & \textbf{0.0} & \textbf{0.0} & \textbf{4.29E-02} & \textbf{0.9} & \textbf{0.0} & \textbf{8.58E-02} & \textbf{0.0} & \textbf{0.0} \\
gplearn / High & 189.1 & 1.12E-01 & 0.0 & \textbf{0.0} & 6.73E-02 & 0.0 & \textbf{0.0} & 1.92E-02 & \textbf{0.0} & \textbf{0.0} & 5.31E-02 & 0.0 & \textbf{0.0} & 3.77E-01 & \textbf{0.0} & \textbf{0.0} \\
\bottomrule
\end{tabular}%
}
\end{table*}

The Low+PostDiag and High comparisons show that diagnosis and additional search are complementary rather than interchangeable. Direct LLM+PostDiag under Qwen3.5-27B exceeds High in Overall accuracy at similar runtime, whereas under Qwen3.5-122B-A10B it has higher accuracy but worse NMSE, exposing a metric tradeoff. For SGA under both backbones, Low+PostDiag approaches the High setting's quality with substantially less end-to-end time, while High retains higher accuracy and lower NMSE. The remaining solvers show similarly method-dependent balances between diagnostic overhead and the benefit of a larger native search budget.
\subsection{Supplementary Results with Additional LLM Backbones}
\label{app:backbone-supplement}

To examine PostDiag under different backbones, we additionally evaluate
Qwen3.5-9B and gpt-oss-20b under the same low-budget protocol.
Table~\ref{tab:backbone-supplement} reports the ID results for the
LLM-based solvers and their PostDiag-augmented variants. All comparisons
within a backbone use the same tasks, random seeds, search budgets,
coefficient-fitting procedure, and validation protocol.

{\color{black}
\begin{center}
\captionof{table}{Supplementary low-budget ID results for Qwen3.5-9B and gpt-oss-20b. Accuracy to tolerance is the primary criterion; NMSE is a tail-sensitive complementary aggregate. Time is the mean end-to-end wall-clock seconds per task run and is reported only in Overall. Metrics use the same three-seed aggregation as Table~\ref{tab:main}.}
\label{tab:backbone-supplement}
\scriptsize
\setlength{\tabcolsep}{1.25pt}
\renewcommand{\arraystretch}{1.08}
\resizebox{\textwidth}{!}{%
\begin{tabular}{llcccccccccccccccc}
\toprule
\multirow{2}{*}{\textbf{Category}} & \multirow{2}{*}{\textbf{Method / setting}} & \multicolumn{4}{c}{\textbf{Overall}} & \multicolumn{3}{c}{\textbf{Phys}} & \multicolumn{3}{c}{\textbf{MatSci}} & \multicolumn{3}{c}{\textbf{Chem}} & \multicolumn{3}{c}{\textbf{Bio}} \\
\cmidrule(lr){3-6}\cmidrule(lr){7-9}\cmidrule(lr){10-12}\cmidrule(lr){13-15}\cmidrule(lr){16-18}
& & Time (s) & NMSE & $A_{.01}$ & $A_{.001}$ & NMSE & $A_{.01}$ & $A_{.001}$ & NMSE & $A_{.01}$ & $A_{.001}$ & NMSE & $A_{.01}$ & $A_{.001}$ & NMSE & $A_{.01}$ & $A_{.001}$ \\
\midrule
\multicolumn{18}{c}{\textit{Qwen3.5-9B}} \\
\midrule
\multirow{6}{*}{LLM-based} & LLM-SR & 129.8 & 2.36E-03 & 14.0 & 5.7 & 5.32E-03 & 9.8 & 3.0 & 2.38E-05 & 37.3 & 17.3 & 4.96E-04 & 6.5 & 2.8 & 2.14E-03 & 8.3 & 2.8 \\
 & \textbf{LLM-SR+PostDiag} & 146.8 & \textbf{3.93E-04} & \textbf{24.0} & \textbf{10.3} & \textbf{7.32E-04} & \textbf{18.2} & \textbf{6.8} & \textbf{1.20E-05} & \textbf{46.7} & \textbf{25.3} & \textbf{7.15E-05} & \textbf{22.2} & \textbf{8.3} & \textbf{6.52E-04} & \textbf{13.9} & \textbf{4.2} \\
\cmidrule(lr){2-18}
 & LaSR & 382.5 & 2.16E-02 & 10.1 & 6.5 & 2.24E-02 & 6.8 & 4.5 & 4.15E-03 & 14.7 & 9.3 & 9.40E-03 & 13.0 & 9.3 & 5.64E-02 & 6.9 & 2.8 \\
 & \textbf{LaSR+PostDiag} & 407.9 & \textbf{3.27E-03} & \textbf{18.6} & \textbf{10.3} & \textbf{6.01E-03} & \textbf{13.6} & \textbf{6.1} & \textbf{6.26E-05} & \textbf{36.0} & \textbf{18.7} & \textbf{8.86E-04} & \textbf{18.5} & \textbf{12.0} & \textbf{5.16E-03} & \textbf{9.7} & \textbf{6.9} \\
\cmidrule(lr){2-18}
 & SR-Scientist & 439.2 & 2.00E-03 & 13.7 & 4.1 & 4.48E-03 & 10.6 & 4.5 & 6.31E-05 & 33.3 & 10.7 & 4.12E-04 & 7.4 & 1.9 & 1.86E-03 & 8.3 & 0.0 \\
 & \textbf{SR-Scientist+PostDiag} & 477.2 & \textbf{5.39E-04} & \textbf{22.5} & \textbf{8.8} & \textbf{9.84E-04} & \textbf{15.9} & \textbf{6.1} & \textbf{1.69E-05} & \textbf{44.0} & \textbf{20.0} & \textbf{1.33E-04} & \textbf{21.3} & \textbf{7.4} & \textbf{8.76E-04} & \textbf{13.9} & \textbf{4.2} \\
\midrule
\multicolumn{18}{c}{\textit{gpt-oss-20b}} \\
\midrule
\multirow{6}{*}{LLM-based} & LLM-SR & 244.7 & 1.31E-03 & 26.1 & 9.6 & 3.18E-03 & 18.9 & 4.5 & 2.62E-05 & 52.0 & 33.3 & 2.03E-04 & 22.2 & 3.7 & 8.91E-04 & 18.1 & 2.8 \\
 & \textbf{LLM-SR+PostDiag} & 260.9 & \textbf{2.39E-04} & \textbf{35.4} & \textbf{15.2} & \textbf{4.31E-04} & \textbf{25.8} & \textbf{9.1} & \textbf{8.52E-06} & \textbf{68.0} & \textbf{45.3} & \textbf{6.47E-05} & \textbf{31.5} & \textbf{8.3} & \textbf{3.87E-04} & \textbf{25.0} & \textbf{5.6} \\
\cmidrule(lr){2-18}
 & LaSR & 395.4 & 1.21E-02 & 10.6 & 5.4 & 1.92E-02 & 9.1 & 5.3 & 3.18E-03 & 13.3 & 5.3 & 8.20E-03 & 13.0 & 7.4 & 1.43E-02 & 6.9 & 2.8 \\
 & \textbf{LaSR+PostDiag} & 419.6 & \textbf{9.53E-04} & \textbf{16.8} & \textbf{9.6} & \textbf{1.71E-03} & \textbf{14.4} & \textbf{8.3} & \textbf{5.04E-05} & \textbf{18.7} & \textbf{10.7} & \textbf{6.83E-04} & \textbf{19.4} & \textbf{11.1} & \textbf{9.13E-04} & \textbf{15.3} & \textbf{8.3} \\
\cmidrule(lr){2-18}
 & SR-Scientist & 134.9 & 8.89E-04 & 33.3 & 15.5 & 2.12E-03 & 24.2 & 8.3 & 3.82E-05 & 53.3 & 29.3 & 1.31E-04 & 38.0 & 17.6 & 6.53E-04 & 22.2 & 11.1 \\
 & \textbf{SR-Scientist+PostDiag} & 159.1 & \textbf{1.60E-04} & \textbf{41.3} & \textbf{23.3} & \textbf{2.68E-04} & \textbf{30.3} & \textbf{17.4} & \textbf{9.47E-06} & \textbf{62.7} & \textbf{37.3} & \textbf{5.03E-05} & \textbf{45.4} & \textbf{24.1} & \textbf{2.81E-04} & \textbf{33.3} & \textbf{18.1} \\
\bottomrule
\end{tabular}%
}
\end{center}
}

\section{PostDiag Core Editor Prompts}
\label{app:editor_prompts}

\subsection{System Prompt}
\begin{PostDiagPrompt}[title after break={System prompt (continued)}]
You are a scientist working on symbolic regression. Use risk localization in the Diagnostic Risk Plane and the corresponding diagnostic evidence map to modify the symbolic structure of a candidate equation. Return a preserved or revised equation skeleton.

The input contains the current candidate, diagnostic coordinates, a diagnostic state, and local behavioral evidence. The numerical procedure has already determined the state. Follow its intervention:

- Settled (sufficiently expressive and stable)
  Prefer preserving the current skeleton. Make only small revisions clearly supported by local evidence; avoid unnecessary terms or replacing the main structure.

- Underfit (insufficiently expressive but stable)
  Preserve the main functional form. Check for missing terms, incorrect operators, or unsuitable variable combinations, and make local repairs targeting the main behavioral discrepancies.

- Overfit (sufficiently expressive but unstable)
  Prefer simplifying the skeleton. Check for redundant terms, repeated subexpressions, complex nesting, and structures that may depend on compensating coefficients.

- Mismatched (insufficiently expressive and unstable)
  You may replace the main functional form. Propose a new structure based on local evidence; unsupported terms or combinations need not be retained.

Local evidence describes differences in level, trend, and roughness between the candidate and observations, as well as behavioral changes after bootstrap refitting of the same skeleton. Focus edits on the structural problems indicated by this evidence. Discrepancy magnitudes do not indicate overprediction or underprediction. Roughness differences may also arise from local nonlinearity; do not treat them as noise alone.

Represent fitted constants as params[i]. An external numerical procedure will fit their values after you return the equation. Small integer exponents may be written as literal constants.

Use only the supplied variables, operators, and functions. Respect the expression-size and nesting constraints. Return one candidate, without loops, conditionals, imports, or additional helper functions. If the evidence does not justify a change, return the original skeleton.

Output only the Python function in the specified format, without explanation. Subsequent numerical fitting and validation determine whether the revision is accepted.
\end{PostDiagPrompt}
\par
\begingroup
\captionsetup{type=figure,skip=6pt}
\captionof{figure}{System prompt for the PostDiag core editor.}
\label{fig:editor-system-prompt}
\endgroup

\subsection{User Prompt Template}
\begin{PostDiagPrompt}[]
# Current candidate equation
{candidate_function}

# Diagnostic results
Skeleton Expressiveness Risk (Bias): {bias}
Skeleton Stability Risk (Var): {variance}
Diagnostic state: {diagnostic_state}

# Local behavioral evidence
The following evidence is computed on the same local patches of the audit set:
- Level discrepancy: the absolute difference between the local means of candidate predictions and observations.
- Trend discrepancy: the norm of the difference between their local slope vectors.
- Roughness discrepancy: the absolute difference between their deviations from their respective local linear approximations.
- Instability: local behavioral variation after bootstrap refitting of the same skeleton. Larger values indicate greater instability.

Patch ID | Level | Trend | Roughness | Instability
{patch_evidence_rows}

# Expression constraints
Input variables: {variables}
Allowed operators and functions: {allowed_operators}
Fitted parameter slots: {parameter_slots}
Expression-size and nesting limits: {structural_constraints}
Input variable ranges: {input_ranges}

# Output requirements
Use the diagnostic state and local evidence to return one preserved or revised equation. Represent every fitted constant as params[i] and use only the supplied parameter slots.

Output a single Python code block with the following function signature. The function body must contain only one return expression:
```python
def equation({signature}, params):
    return ...
```
\end{PostDiagPrompt}
\par
\begingroup
\captionsetup{type=figure,skip=6pt}
\captionof{figure}{User prompt template for the PostDiag core editor.}
\label{fig:editor-user-prompt}
\endgroup

\end{document}